\documentclass{article}

\usepackage{booktabs}

\usepackage[final]{neurips_2025}

\usepackage{graphicx}
\usepackage[utf8]{inputenc} 
\usepackage[T1]{fontenc}    
\usepackage{hyperref}       
\usepackage{url}            
\usepackage{amsfonts}       
\usepackage{nicefrac}       
\usepackage{microtype}      
\usepackage{xcolor}         
\usepackage{xspace}
\usepackage{adjustbox}
\usepackage{subcaption}

\usepackage{multirow}
\usepackage{amsmath}
\usepackage{amssymb}
\usepackage{mathtools}
\usepackage{amsthm}
\usepackage{algorithm}
\usepackage{algorithmic}
\usepackage{bm}
\usepackage[capitalize,noabbrev]{cleveref}
\usepackage{wrapfig}
\theoremstyle{plain}
\newtheorem{theorem}{Theorem}[section]
\newtheorem{proposition}{Proposition}[section]

\theoremstyle{definition}

\newtheorem{assumption}{Assumption}[section]
\theoremstyle{remark}

\DeclareMathOperator{\Var}{Var} 

\title{Aligning by Misaligning: Boundary-aware Curriculum Learning for Multimodal Alignment}

\author{
  Hua Ye\textsuperscript{1,2*},
  Hang Ding\textsuperscript{3*},
  Siyuan Chen\textsuperscript{4},
  Yiyang Jiang\textsuperscript{5},
  Changyuan Zhang\textsuperscript{6},
  Xuan Zhang\textsuperscript{2,7\textdagger}
  \\
  \small
  \textsuperscript{1}Nanjing University \quad
  \textsuperscript{2}Airon Technology CO., LTD \quad
  \textsuperscript{3}Shanghai Jiao Tong University \\
  \small
  \textsuperscript{4}University of Bristol \quad
  \textsuperscript{5}The Hong Kong Polytechnic University \\
  \small
  \textsuperscript{6}The University of Hong Kong \quad
  \textsuperscript{7}Carnegie Mellon University
}

\makeatletter
\let\oldmaketitle\maketitle
\renewcommand{\maketitle}{%
  \oldmaketitle%
  \thispagestyle{plain}%
  \begingroup
  \renewcommand{\thefootnote}{}%
  \footnotetext{* Equal contribution, \textdagger Corresponding author(xuanzhang2199@gmail.com)}%
  \endgroup
}
\makeatother

\begin{document}

\maketitle

\begin{abstract}
Most multimodal models treat every negative pair alike, ignoring the
ambiguous negatives that differ from the positive by only a
small detail.  We propose Boundary-A ware Curriculum with Local Attention(BACL), a lightweight add-on that
turns these borderline cases into a curriculum signal.  A
Boundary-aware Negative Sampler gradually raises difficulty,
while a Contrastive Local Attention loss highlights where the
mismatch occurs.  The two modules are fully differentiable and work
with any off-the-shelf dual encoder.  Theory predicts a fast
$\tilde{\mathcal{O}}(1/n)$ error rate; practice shows up to
+32 \% R@1 over CLIP and new SOTA on four large-scale
benchmarks, all without extra labels.
\end{abstract}

\section{Introduction}
\label{sec:intro}
Cross–modal representation learning has witnessed rapid progress since
CLIP~\citep{clip2021}, ALIGN~\citep{jia2021align} and their successors
demonstrated that contrastive pre-training on web-scale
image–text pairs is an effective alternative to costly human
annotation~\citep{li2025mmt}.  
Follow-up models such as ALBEF~\citep{albef2021},
BLIP/BLIP-2~\citep{li2022blip,li2023blip2},
ViLT~\citep{kim2021vilt} and GRAM~\citep{cicchetti2024gram} further
increase sample efficiency by injecting token-level objectives or
multi-modal experts\citep{zhang2025enhancing,zhang2024cf}. These models have promoted the development of fields such as natural language processing~\citep{lin2025cec}, medical diagnosis~\citep{fang2025regionmed,liu2025lightweightbaselinesmedicalabstract,tong2025renaissance,wang2025medical,Li2025Efficient}, autonomous systems\citep{yao2023ndc,lu2025clip,xiao2025diffusion,lu2024drivingrecon,li2025slam,zeng2025FSDrive,zeng2025janusvln}, and other application fields~\citep{xu2024fakeshield,jiang2025transforming,tao2023dudb,liao2025convex,chan2026adagar,10.1145/3711896.3737195,wang-etal-2025-reasoning-enhanced}.

Despite these advances, most existing pipelines share the \emph{same
implicit assumption}: two paired modalities are either perfectly
aligned (\textit{positive}) or entirely unrelated (\textit{negative}),
and the learner’s job is merely to shorten or enlarge their distance~\citep{xin2024mmap,fu2024touch,tan2025profix,yu2025pm}.
In practical applications\citep{10.1145/3664647.3681115,zhang2023multi}, however, cross-modal data often carries
subtle mismatches: captions that paraphrase only part of an
image, audio tracks that overlap but differ in background context,
or video–subtitle pairs where a single phrase is out of sync.

These \emph{ambiguous negatives}---\textquotedblleft
half-true, half-false\textquotedblright\ mismatches---are abundant
on the web yet overlooked by current training regimes~\citep{yang2024trisampler,yao2024swift}.Humans, by contrast, naturally learn from nuanced differences, readily using subtle mismatches as informative cues~\citep{osgood1949similarity,lynn2020humans,hu-etal-2025-removal,sun2025objective,MAIL}. Enabling models to similarly leverage these ambiguous negatives is thus essential for more human-like multimodal alignment.

\begin{figure}[t]
    \centering
    \includegraphics[width=0.9\linewidth]{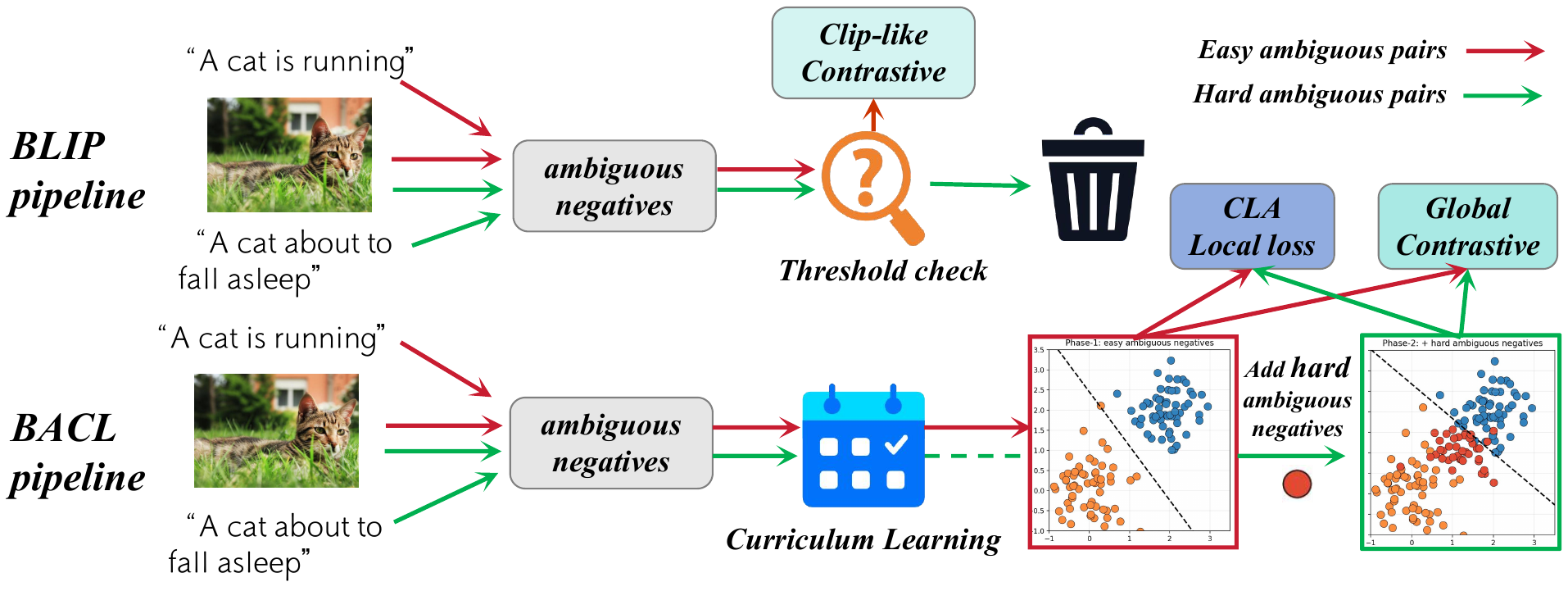} 
    \caption{Comparison between the BLIP pipeline and our proposed BACL pipeline. Methods like BLIP eliminate ambiguous negatives through threshold filtering without explicitly leveraging their intrinsic value. In contrast, BACL employs a curriculum learning strategy to progressively introduce more challenging ambiguous negative samples, explicitly revealing the sources of confusion. This approach enhances discriminative capability by jointly optimizing the global contrastive loss and the Contrastive Local Attention (CLA) loss.}
    \label{fig:teaser}
\end{figure}

\paragraph{Gap in current practice.}
Existing alignment methods inadequately handle ambiguous negatives. 
\emph{First}, mainstream dual-encoder approaches (e.g., CLIP~\citep{clip2021}, ALIGN~\citep{jia2021align}, MIL-NCE~\citep{miech2020end}) sample negatives uniformly, treating obvious mismatches and subtly incorrect captions equally. 
\emph{Second}, recent token-aware methods (e.g., ALBEF~\citep{albef2021}, BLIP~\citep{li2022blip}) discard ambiguous captions through filtering or pseudo-labeling, losing valuable instructional signals, as shown in Fig ~\ref{fig:teaser}. 
\emph{Finally}, existing approaches rely exclusively on static datasets and static loss functions, neglecting dynamically generated, structurally plausible yet semantically ambiguous mismatches~\citep{yan2024sard,xiao2025multifrequency}. Consequently, these methods optimize alignment under idealized conditions, overlooking rich supervisory signals inherent in realistic, partially incorrect data~\citep{wang2024computing,lu2025dataset}.

\paragraph{Our view.}
We argue that ambiguous negatives are not merely \emph{noise},
but a rich supervisory signal: distinguishing \emph{almost-correct}
pairs from truly correct ones is precisely what a robust multimodal
model must master when data are scarce or noisy.
However, directly training on these borderline cases from the
beginning leads to unstable optimisation~\citep{comm2024}.
The key, therefore, is to \emph{schedule} the exposure of the learner
to increasingly confusing negatives while simultaneously revealing
\emph{where} the confusion arises.

\paragraph{Our solution.}
This paper introduces \textbf{BACL}:
\underline{B}oundary-\underline{A}ware \underline{C}urriculum with
\underline{L}ocal Attention. BACL augments any dual-encoder or MoE aligner with two lightweight,
fully differentiable components:
(i)~\emph{Boundary-aware Negative Sampler (BNS).}  
        A policy network learns to rank candidate negatives by their
        boundary score and, guided by a logistic schedule,
        gradually shifts training focus from easy to ambiguous cases.
(ii)~\emph{Contrastive Local Attention (CLA).} 
        For every positive pair, CLA compares its cross-attention map
        with that of the hardest negative and imposes a local mismatch
        loss that amplifies token pairs where they diverge, forcing the
        encoder to detect fine-grained misalignment.

\paragraph{Contributions.}
The main contributions are:

1) We identify \emph{ambiguous negatives} as an under-explored yet
        ubiquitous phenomenon in web-scale multimodal corpora and
        highlight their importance for robust alignment.
We propose BACL, the first framework that \emph{dynamically}
        generates and exploits near-boundary mismatches via a
        curriculum sampler and a token-level attention loss.

2) We provide a \emph{sharp generalisation theory}:
        under mild assumptions BACL enjoys a \(\tilde{\mathcal{O}}(1/n)\)
        \emph{fast rate}, whereas uniform sampling suffers an
        unavoidable \(\Omega(\rho/\sqrt{n})\) excess risk
        (Theorems~\ref{thm:upper-main}–\ref{thm:lower-main}).

3) Extensive experiments on four large-scale datasets show that BACL improves both global retrieval and fine-grained reasoning, achieving new
        state-of-the-art results on several benchmarks.
\section{Related Works}
\label{sec:related}

\paragraph{Multimodal Alignment and Hard-Negative Mining.}
Early dual-encoder models such as CLIP~\citep{clip2021},
ALIGN~\citep{jia2021align} and MIL-NCE~\citep{miech2020milnce} learn a
shared representation space by drawing \emph{uniform} in-batch
negatives, an approach that proves scalable but treats trivial and
near-boundary mismatches alike~\citep{dong2024pixel}.
Subsequent work improves sample efficiency through stronger vision
backbones~\citep{jia2021align}, region-level supervision~\citep{uniter2020},
or generative pre-text objectives
(ALBEF~\citep{albef2021}, BLIP/BLIP-2~\citep{li2022blip,li2023blip2}).
Yet, these methods either filter out noisy captions
(ALBEF, BLIP) or rely on one-shot
max-violation mining (VSE++~\citep{faghri2018vsepp}), leaving
\textit{ambiguous negatives} under-exploited.
Very recent systems such as GRAM~\citep{cicchetti2024gram} and
Emergence~\citep{emergence2024} incorporate mixture-of-experts or
multi-modal masking but still assume a static positive/negative split\citep{xiao2025curiosity}.
Our work departs from this line by \emph{dynamically} scheduling
near-boundary negatives and coupling them with a local attention loss,
thereby tightening the decision margin without expensive expert routing
or additional labels.

\paragraph{Curriculum and Self-Paced Learning.}
Curriculum learning~\citep{bengio2009curriculum} and
self-paced learning~\citep{kumar2010self,xiao2024confusion} propose to present
samples in an easy-to-hard order to stabilise optimisation.
In computer vision, curricula have been applied to
class imbalance~\citep{jiang2015self},
structured prediction~\citep{pentina2015curriculum},
and more recently to vision-language pre-training
DCOT~\citep{dcot2023},
where difficulty is measured heuristically (e.g.\ OT distance).
Our \emph{Boundary-aware Negative Sampler} differs in two aspects:
(i)~difficulty is defined by a \emph{learnable boundary score}
relative to the current decision margin, and
(ii)~sampling remains \emph{differentiable} via Gumbel–Softmax,
enabling end-to-end optimisation with the underlying encoders.
BACL constitutes the first
curriculum framework specifically designed to exploit
\textquotedblleft half-true, half-false\textquotedblright\ negatives in
multimodal alignment.
\section{Methods}
\label{sec:method}
\subsection{Overview of the Proposed Method}
\label{subsec:overview}
\textbf{Problem Definition.}
Consider two arbitrary modalities $\mathcal{X}$ and $\mathcal{Y}$  
(e.g., image–text, audio–video).  
A large paired corpus is denoted by  
$\mathcal{D}=\{(x_i,y_i)\}_{i=1}^{N}$ with modality--specific encoders  
$\phi_{\mathcal{X}}$ and $\phi_{\mathcal{Y}}$ that map inputs to a
shared unit sphere $\mathbb{S}^{d-1}$.  
For a ground–truth pair $(x,y)$ we measure similarity by
$s(x,y)=\langle\phi_{\mathcal{X}}(x),\phi_{\mathcal{Y}}(y)\rangle$.  
A non-matched sample $z$ is called an \emph{ambiguous negative} if
\begin{equation}
\label{eq:and}
\bigl|\,s(x,z)-s(x,y)\bigr|\le\varepsilon,
\quad\text{with }\;0<\varepsilon\ll1,
\end{equation}
i.e., it lies near the decision boundary and is difficult to reject.
Standard alignment strategies treat all negatives uniformly and thus fail
to exploit this hardest subset.

\textbf{Core Idea.}
We explicitly expose the model to such boundary cases through a
\emph{boundary-aware curriculum}:
1) \emph{Boundary-aware Negative Sampler (BNS)}
      progresses from easy to ambiguous negatives.
      A scheduling coefficient $\alpha(\eta)$, monotonically increasing
      with epoch~$\eta$, controls the hardness of the sampled set,
      thereby shrinking the margin around every positive pair.
2) \emph{Contrastive Local Attention (CLA)}  
      contrasts the attention patterns of a positive pair
      with those of its hardest negative counterpart selected by BNS.
      The resulting local mismatch loss encourages the model to
      pinpoint fine-grained misalignment cues rather than relying solely
      on global similarity.

By iteratively tightening the boundary (via BNS) and highlighting
where misalignment occurs (via CLA), the framework learns
finer-grained and more robust cross-modal representations.


\subsection{Boundary-aware Negative Sampler}
\label{subsec:bns}

In latent space, negatives that closely resemble a positive pair are the hardest to reject.
We introduce a learnable \emph{Boundary-aware Negative Sampler} (BNS) $\pi_\theta$ that schedules negatives from easy to hard, steadily shifting training focus toward these ambiguous cases and, in turn, sharpening the decision boundary.

\paragraph{Boundary Score and Candidate Negatives}
First, using an \emph{initially trained coarse alignment model}, we encode all images $I$ and texts $T$ in the training set into embeddings $\mathbf{z}_{(I)}$ and $\mathbf{z}_{(T)}$, respectively, and build corresponding image and text indices. For any given positive pair $(I, T)$, we retrieve texts $\{T'\}$ from the text index that are most similar to $\mathbf{z}_{(I)}$ yet do not form true positive pairs. Similarly, we retrieve images $\{I'\}$ from the image index that closely match $\mathbf{z}_{(T)}$ but are not true matches. We collectively refer to these retrieved samples as \emph{ambiguous negative samples}, and denote their embeddings as $\{\mathbf{z}_n\}_{n=1}^N$.

To quantify how challenging each candidate negative is with respect to the model’s decision boundary, we define a \emph{boundary score} function:
\begin{equation}
  \label{eq:bscore}
  \mathrm{BS}\bigl(\mathbf{z}_{(I)}, \mathbf{z}_{(T')}\bigr)
  \;=\;
  \mathrm{sim}\bigl(\mathbf{z}_{(I)}, \mathbf{z}_{(T')}\bigr)
  \;-\;
  \mathrm{sim}\bigl(\mathbf{z}_{(I)}, \mathbf{z}_{(T)}\bigr),
\end{equation}
where $\mathrm{sim}(\cdot,\cdot)$ denotes cosine similarity. A large positive boundary score indicates that the negative sample $T'$ is even closer to image $I$ than the true matching text $T$, presenting a substantial confusion to the model. A score close to zero implies that the negative sample is almost indistinguishable from the true pair, thus also posing significant confusion. Conversely, a clearly negative score suggests the negative sample is relatively distant from $I$ in the embedding space and poses a weaker challenge to the model. The same boundary score definition applies similarly to the retrieved pairs $(I, T')$ or $(I', T)$.

\paragraph{Policy Network and Difficulty Scheduling}
We define a policy network $\pi_\theta(\cdot)$, which takes as input the embeddings of the current positive pair $(\mathbf{z}_{(I)}, \mathbf{z}_{(T)})$ and all candidate negative samples $\{\mathbf{z}_n\}_{n=1}^N$, outputting an initial scoring vector $(u_1,\ldots,u_N)$, where $u_n$ quantifies the priority of selecting each negative sample. Next, we introduce the following difficulty measure:
\begin{equation}
  d(\mathbf{z}_n)
  \;=\;
  \max\Bigl\{0,\;\mathrm{sim}\bigl(\mathbf{z}_{(I)}, \mathbf{z}_n\bigr)
               \;-\;\mathrm{sim}\bigl(\mathbf{z}_{(I)}, \mathbf{z}_{(T)}\bigr)\Bigr\}.
\end{equation}
A larger value of $d(\mathbf{z}_n)$ indicates that the negative sample poses a greater confusion to the model. To balance the principle of progressing from easier to more challenging negatives, we compute the difficulty measure $d(\mathbf{z}_n)$ for each negative sample and adjust the initial scores accordingly using a function $\alpha(\eta)$:
\begin{equation}
  \label{eq:hat_un}
  \hat{u}_n 
  \;=\;
  u_n \;-\; \alpha(\eta)\,\,d(\mathbf{z}_n),
\end{equation}
where $\alpha(\eta)$ adopts the form of a \emph{logistic function} that smoothly transitions from $\alpha_{\text{early}}>0$ to $\alpha_{\text{late}}<0$:
\begin{equation}
  \label{eq:alpha_eta}
  \alpha(\eta)
  \;=\;
  \alpha_{\text{early}}
  \;+\;
  \Bigl(\alpha_{\text{late}} \;-\; \alpha_{\text{early}}\Bigr)
  \;\frac{1}{1 + \exp\bigl(-\,\gamma\,(\,\eta - \eta_0)\bigr)}.
\end{equation}
Here, $\eta$ denotes the training epoch, $\gamma>0$ controls the steepness of the transition, and $\eta_0$ represents the center point of the transition. At early stages of training ($\eta \ll \eta_0$), $\alpha(\eta)$ remains close to $\alpha_{\text{early}}>0$, thereby suppressing highly challenging negative pairs. Conversely, at later stages ($\eta \gg \eta_0$), $\alpha(\eta)$ approaches $\alpha_{\text{late}}<0$, thus incentivizing the sampling of more confusing negative samples.

\paragraph{Differentiable Sampling Mechanism}
The adjusted scores $\hat{u}_n$ are transformed into probabilities $\tilde{p}_n$ via the Gumbel-Softmax operation:
\begin{equation}
  \tilde{p}_n
  \;=\;
  \frac{\exp\bigl((\hat{u}_n + g_n)/\tau\bigr)}
       {\sum_{m=1}^N \exp\bigl((\hat{u}_m + g_m)/\tau\bigr)},
\end{equation}
where $g_n$ denotes Gumbel noise, $\tau$ is a temperature hyperparameter, and $\tilde{p}_n$ approximates the probability of sampling negative sample $\mathbf{z}_n$.

\paragraph{Upper-level Optimization Objective}
We treat the boundary score defined in Eq.~\eqref{eq:bscore} as a reward signal $R(\mathbf{z}_n)$ and define:
\begin{equation}
  \label{eq:sampler_objective}
  J(\theta)
  \;=\;
  \mathbb{E}_{\mathbf{z}_n\sim \pi_\theta}\bigl[R(\mathbf{z}_n)\bigr].
\end{equation}
Thanks to the differentiable Gumbel-Softmax mechanism, we can directly perform backpropagation on the following expectation to update the policy network parameters $\theta$:
\begin{equation}
  \sum_{n=1}^N \tilde{p}_n \, R(\mathbf{z}_n).
\end{equation}
Early on, BNS down-weights the hardest negatives so the model can master basic discrimination; as training proceeds, it gradually upsamples the most ambiguous cases, tightening the margin and yielding finer cross-modal distinctions.

\subsection{Contrastive Local Attention}
\label{subsec:local_attention}

Global contrastive loss separates pairs overall but misses token-level mismatches.
Our \emph{Contrastive Local Attention} (CLA) compares the attention maps of a positive pair with its hardest negative, amplifying the tokens where they diverge and spotlighting fine-grained misalignments.

\paragraph{Attention Distributions of Positive and Negative Pairs}
Within the cross-modal Transformer, let the attention matrix for a positive pair $(I, T)$ be denoted by $\mathbf{A}^{(+)} \in \mathbb{R}^{N \times N}$, where $N = M + L$ represents the total number of image and text tokens. If the sampler selects a challenging negative pair $(I, T')$ (or $(I', T)$) corresponding to the positive pair $(I, T)$, we similarly obtain the negative-pair attention matrix $\mathbf{A}^{(-)}$. These matrices respectively reflect the differences in attention distributions across tokens between the positive and negative pair scenarios.

\paragraph{Difference Computation and Local Modulation}
When a negative pair $(I,T')$ significantly differs from the corresponding positive pair at certain token pairs $(i,j)$, these positions usually indicate potential mismatches. To amplify such differences, we define:
\begin{equation}
  \label{eq:delta_attn}
  \mathbf{\Delta A}(i,j)
  \;=\;
  \bigl|\mathbf{A}^{(+)}(i,j)
         \;-\;
         \mathbf{A}^{(-)}(i,j)\bigr|.
\end{equation}
A higher value of $\mathbf{\Delta A}(i,j)$ suggests a substantial discrepancy between positive and negative pairs at token pair $(i,j)$, typically corresponding to regions of highest ambiguity in negative samples. To direct the model’s attention more explicitly toward these critical regions in negative-pair scenarios, we locally enhance the negative attention matrix $\mathbf{A}^{(-)}$ as follows:
\begin{equation}
  \label{eq:cla_attention}
  \mathbf{A}^{b}(i,j)
  \;=\;
  \mathbf{A}^{(-)}(i,j)
  \;\times\;
  \bigl[\,1 \;+\; \beta\,\mathbf{\Delta A}(i,j)\bigr],
\end{equation}
where $\beta>0$ denotes a gain coefficient that emphasizes token pairs exhibiting large attention discrepancies.

\paragraph{Local Mismatch Loss}
After obtaining the modulated attention matrix $\mathbf{A}^{b}$, we introduce a local mismatch loss $\mathcal{L}_{\mathrm{local}}$ to further emphasize these mismatched regions, defined as:
\begin{equation}
  \label{eq:local_loss_cla}
  \mathcal{L}_{\mathrm{local}} 
  \;=\;
  \sum_{(i,j)\,\in\,\Omega}
  g\bigl(\mathbf{A}^{b}(i,j)\bigr),
\end{equation}
where $\Omega$ represents a set of token pairs with the highest discrepancies identified by Eq.~\eqref{eq:delta_attn} (selected via thresholding or ranking). The function $g(\cdot)$ can be instantiated as $-\log(\cdot)$, thereby encouraging the model to produce a more pronounced attention enhancement at potential mismatch locations.

We combine the aforementioned local mismatch loss $\mathcal{L}_{\mathrm{local}}$ with the global contrastive loss $\mathcal{L}_{\mathrm{contrast}}$ to form the final training objective:
\begin{equation}
  \label{eq:final_loss}
  \mathcal{L}_{\mathrm{main}}
  \;=\;
  \mathcal{L}_{\mathrm{contrast}}
  \;+\;
  \lambda_{\mathrm{local}} \,\mathcal{L}_{\mathrm{local}},
\end{equation}
The weight $\lambda_{\mathrm{local}}$ trades off global contrast with token-level mismatch loss.
By amplifying attention gaps between a positive pair and its hardest negative, CLA trains the model to catch both pair-wise and token-wise errors, sharpening its response to ambiguous negatives. Experiments (§\ref{sec:exp}) show that CLA, combined with BNS, boosts accuracy and fine-grained alignment.

Algorithm \ref{alg:bacl} in Appendix \ref{app:AL} outlines our \textsc{BACL} training pipeline.
The sampler–attention synergy progressively exposes the encoders to increasingly ambiguous negatives while amplifying token-level mismatch cues, yielding finer cross-modal decision boundaries.

\section{Theoretical Analysis}
\label{sec:theory}
We investigate the sample–complexity and optimisation behaviour of
\textsc{BACL}.  Throughout, Assumptions~\ref{asm:a1}–\ref{asm:a2} hold;
all proofs are deferred to Appendix~\ref{app:proofs}.

\begin{assumption}[Ambiguous–negative density]\label{asm:a1}
There exists $\rho\in(0,1)$ such that for every anchor $x$
and its positive $y^{+}$,
\(
  \Pr_{z\sim\mathcal{Y}}\bigl(|s(x,z)-s(x,y^{+})|\le\varepsilon\bigr)
  =\rho ,
\)
where $0<\varepsilon\ll m$.
\end{assumption}

\begin{assumption}[Lipschitz encoders]\label{asm:a2}
The encoders are $L$–Lipschitz:
\(
  |s(\phi_{\mathcal{X}}(x),\phi_{\mathcal{Y}}(y))
    -s(\phi_{\mathcal{X}}(x'),\phi_{\mathcal{Y}}(y'))|
  \le
  L\bigl(\|x-x'\|+\|y-y'\|\bigr)
\)
for all $x,x',y,y'$.
\end{assumption}

\noindent\textbf{Notation.}
Let $\hat{\phi}_{\textsc{B}}$ be the model returned by
Algorithm~\ref{alg:bacl};
$\hat{\phi}_{\textsc{U}}$ denotes the counterpart trained with \emph{uniform} negatives.
The population risk is
\(
\mathcal{R}(\phi)=
\mathbb{E}\bigl[\mathcal{L}_{m}(x,y^{+},z;\phi)\bigr].
\)

\begin{theorem}[Fast–rate Generalisation of \textsc{BACL}]
\label{thm:upper-main}
Assume \ref{asm:a1} and \ref{asm:a2}.  Fix $\delta\!\in\!(0,1)$, margin
$m\!>\!\varepsilon$ and let $d_{\text{eff}}$ be the effective (pseudo)
dimension of $\Phi$.  If
\begin{equation}
n
\;\ge\;
\frac{128\,L^{2}}{(1-\rho)^{2}(m-\varepsilon)^{2}}
\Bigl(d_{\text{eff}}
      \,\log\!\frac{4\,e}{m-\varepsilon}
      +\log\!\frac{4}{\delta}
\Bigr),
\end{equation}
then with probability at least $1-\delta$
\begin{equation}
\bigl|
\mathcal{R}(\hat{\phi}_{\textsc{B}})
-
\mathcal{R}(\phi^{\star})
\bigr|
\;\le\;
\frac{16\,L(m-\varepsilon)}{1-\rho}\;
\sqrt{\frac{
        d_{\text{eff}}\log\frac{4e}{m-\varepsilon}+\log\frac{4}{\delta}}
     {n}}
\;+\;
\frac{32\,L^{2}}{(1-\rho)n},
\label{eq:upper-bound}
\end{equation}
where the additional $L^{2}/n$ term refines the classical
$\tilde{\mathcal{O}}(1/\sqrt{n})$ rate to a \emph{fast rate} whenever
$m-\varepsilon=\Theta(1)$.
\end{theorem}

\begin{theorem}[Minimax Lower Bound for Uniform Samplers]
\label{thm:lower-main}
Let Assumptions~\ref{asm:a1}--\ref{asm:a2} hold.
Fix $\rho\!\in\!(0,\tfrac12)$, $\varepsilon\!<\!\tfrac{m}{4}$ and
$\delta\!\in\!(0,\tfrac14)$.  
For any estimator\footnote{%
An \emph{estimator} is any (possibly randomised) measurable mapping
$\Psi:\!\bigl((\mathcal{X}\!\times\!\mathcal{Y}\!\times\!\mathcal{Y})^{n},
\{\text{uniform neg.\ sample}\}\bigr)\!\to\!\Phi$;
$\hat\phi=\Psi(\text{data})$.}  
$\hat{\phi}$ trained with \emph{uniform} negatives on $n$ triplets,
there exists a distribution $\mathbb{P}$ satisfying
Assumptions~\ref{asm:a1}--\ref{asm:a2} such that, with probability at
least $1-2\delta$,
\begin{equation}
\mathcal{R}(\hat\phi)-\mathcal{R}(\phi^{\star})
\;\ge\;
\frac{\rho(m-2\varepsilon)}{32}\;
\sqrt{\frac{\log\!\tfrac{1}{4\delta}}{\,n}}
\;+\;
\frac{\rho^{2}L(m-2\varepsilon)^{2}}{128n}.
\label{eq:lower-bound}
\end{equation}
Hence any learner that \emph{ignores} ambiguous negatives incurs an
\textbf{unavoidable} $\Omega\!\bigl(\rho/\sqrt{n}\bigr)$ excess risk,
matching the fast rate in \ref{thm:upper-main} up to constants.
\end{theorem}

\begin{proposition}[Exponential Contraction of Alignment Margin]
\label{prop:margin-main}
Let $\Delta_{\eta}$ denote the \emph{expected worst–case margin}
after epoch $\eta$,
\(
  \Delta_{\eta}
   \!=\!
   \mathbb{E}\bigl[
     s(x,y^{+})
     -\!\!\!\max_{z\in\mathcal{N}_{\mathrm{hard}}}
           \!s(x,z)
   \bigr].
\)
Assume a constant learning rate $\eta_{\mathrm{lr}}$,
batch size $B$, and $\beta\!\ge\!1$ in CLA.
Define
\(
  \bar{\alpha}_{\eta}
   =\frac{1}{B}\sum_{t=1}^{\eta}\alpha(t)
\)
and let
\(
  \kappa
   =\frac{\eta_{\mathrm{lr}}\beta(m-\varepsilon)}{2L}>0.
\)
Then for all $\eta\!\ge\!1$
\begin{equation}
\Delta_{\eta}
\;\le\;
\Delta_{0}\;
\exp\!\Bigl(
  -\kappa\,
  \bigl(e^{\,\bar{\alpha}_{\eta}}
        -1\bigr)
\Bigr).
\label{eq:margin-super}
\end{equation}
\end{proposition}
Consequently, if the logistic schedule
$\alpha(\eta)$ of \eqref{eq:alpha_eta} obeys
$\alpha_{\text{late}}\!\le\!-c_{\alpha}\!(<\!0)$, then
$\Delta_{\eta}\!=\!\mathcal{O}\!\bigl(
  e^{-\,\Theta(\eta^{2})}
\bigr)$ once $\eta\!>\!\eta_{0}$,
implying that a quadratic number of ambiguous epochs suffices to
force the margin below any preset $\varepsilon_{\text{target}}$.
\section{Experiments}
\label{sec:exp}
\subsection{Experimental Setup}

\textbf{Datasets}  
We evaluate the effectiveness of the proposed \textsc{BACL} framework on four large-scale multimodal datasets that naturally contain
``ambiguous negatives'' (near-boundary hard negatives) across different modality pairs:
(i) the \emph{LAION-400M} image–text corpus~\citep{schuhmann2021laion};
(ii) the \emph{WebVid-10M} video–text collection~\citep{bain2021frozen};
(iii) the \emph{VAST-27M} tri-modal dataset of video, audio, and subtitles~\citep{chen2023vast};
and (iv) the \emph{WavText5K} audio–text benchmark~\citep{deshmukh2022audio}.
For a detailed description of these datasets, including their statistics, licensing, and the
the evaluation metrics we follow, please refer to Appendix~\ref{sec:app_exp_data}.
We additionally report extended results on VQA and NLVR2 to
probe fine-grained reasoning; these results appear in Appendix~\ref{sec:extended_exp}.

\textbf{Baselines}
We benchmark the proposed \textsc{BACL} against five families of
vision–language alignment methods:
(i)~\emph{Uniform-negative dual encoders} 
(CLIP~\citep{clip2021}, ALIGN~\citep{jia2021align}).
(ii)~\emph{Single-shot hard-negative mining}  
        (VSE++~\citep{faghri2018vsepp}, UNITER~\citep{uniter2020},
        ALBEF~\citep{albef2021}).
(iii)~\emph{Token-level enhanced pre-training}  
        (ViLT~\citep{kim2021vilt}, BLIP~\citep{li2022blip},
        BLIP-2~\citep{li2023blip2}).
(iv)~\emph{Curriculum or self-paced alignment}  
        (DCOT~\citep{dcot2023}).
(v)~\emph{Multimodal or MoE aligners}  
    (Emergence~\citep{emergence2024}, CoMM~\citep{comm2024},
        M3-JEPA~\citep{m3jepa2024}, GRAM~\citep{cicchetti2024gram},
        CLAP~\citep{clap2022} for audio, MIL-NCE~\citep{miech2020milnce}
        for video).

Baselines are evaluated using their public checkpoints or our reproduction with recommended hyper-parameters (Appendix~\ref{app:baseline}). Implementation details are in Appendix~\ref{sec:app_exp}.

\subsection{Main Results}\label{sec:main_results}

\begin{table*}[ht]
\centering
\caption{Retrieval performance on \textbf{(a) LAION‑400M} (image–text) and \textbf{(b) WebVid‑10M} (video–text).
Higher is better; best results are in bold.}
\label{tab:retrieval12}
\begin{subtable}[t]{0.48\textwidth}
  \subcaption{LAION‑400M}
  \label{tab:laion}
  \centering
  \resizebox{\linewidth}{!}{%
  \begin{tabular}{lcccc}
    \toprule
    \textbf{Method} & \textbf{R@1} & \textbf{R@5} & \textbf{R@10} & \textbf{mAP} \\
    \midrule
    CLIP~\citep{clip2021}            & 35.2 & 58.3 & 68.7 & 42.3 \\
    ALIGN~\citep{jia2021align}       & 37.9 & 61.5 & 71.2 & 44.6 \\
    VSE++~\citep{faghri2018vsepp}    & 18.4 & 35.7 & 46.1 & 22.5 \\
    UNITER~\citep{uniter2020}        & 32.7 & 56.1 & 66.0 & 38.8 \\
    ALBEF~\citep{albef2021}          & 40.8 & 65.8 & 74.2 & 47.9 \\
    ViLT~\citep{kim2021vilt}         & 32.6 & 55.4 & 64.3 & 39.2 \\
    BLIP~\citep{li2022blip}          & 42.0 & 67.3 & 76.1 & 49.2 \\
    DCOT~\citep{dcot2023}            & 41.1 & 66.4 & 75.3 & 48.7 \\
    Emergence~\citep{emergence2024}  & 38.3 & 63.9 & 73.0 & 45.8 \\
    CoMM~\citep{comm2024}            & 39.2 & 64.5 & 74.0 & 46.4 \\
    M3‑JEPA~\citep{m3jepa2024}       & 43.3 & 68.1 & 76.4 & 50.1 \\
    GRAM~\citep{cicchetti2024gram}   & 44.0 & 69.0 & 77.0 & 50.8 \\
    \midrule
    \textbf{CLIP+BACL (Ours)}        & \textbf{46.5} & \textbf{71.2} & \textbf{79.3} & \textbf{53.6} \\
    \textbf{M3‑JEPA+BACL (Ours)}     & 46.0 & 70.5 & 78.9 & 52.9 \\
    \bottomrule
  \end{tabular}}
\end{subtable}
\hfill
\begin{subtable}[t]{0.48\textwidth}
  \subcaption{WebVid‑10M}
  \label{tab:webvid}
  \centering
  \resizebox{\linewidth}{!}{%
  \begin{tabular}{lcccc}
    \toprule
    \textbf{Method} & \textbf{R@1} & \textbf{R@5} & \textbf{R@10} & \textbf{nDCG} \\
    \midrule
    CLIP~\citep{clip2021}            & 14.3 & 31.5 & 42.7 & 25.4 \\
    ALBEF~\citep{albef2021}          & 15.6 & 34.5 & 45.8 & 26.3 \\
    BLIP~\citep{li2022blip}          & 17.2 & 36.8 & 47.5 & 28.0 \\
    DCOT~\citep{dcot2023}            & 18.0 & 38.1 & 48.4 & 29.1 \\
    MIL‑NCE~\citep{miech2020milnce}  & 12.4 & 28.1 & 38.9 & 22.7 \\
    M3‑JEPA~\citep{m3jepa2024}       & 21.4 & 42.7 & 53.0 & 32.4 \\
    GRAM~\citep{cicchetti2024gram}   & 22.0 & 43.6 & 54.1 & 33.0 \\
    \midrule
    CLIP+BACL (Ours)                 & 19.5 & 35.0 & 51.5 & 31.4 \\
    \textbf{MIL‑NCE+BACL (Ours)}     & \textbf{24.9} & \textbf{46.8} & \textbf{57.3} & \textbf{35.9} \\
    M3‑JEPA+BACL (Ours)              & 23.8 & 45.9 & 56.8 & 35.0 \\
    \bottomrule
  \end{tabular}}
\end{subtable}
\end{table*}

\textbf{Image–text retrieval (Table~\ref{tab:laion}).}
On the noisy web-scale LAION-400M corpus, \textsc{BACL} injects an
\(\mathbf{+32\%}\) relative gain in R@1 over vanilla CLIP and still
offers \(\,{\approx}6\%\) absolute improvement over sophisticated
hard-negative methods such as GRAM.  

\textbf{Video–text retrieval (Table~\ref{tab:webvid}).}
WebVid-10M captions are notoriously weak: many clips share nearly
identical phrases.  Curriculum sampling therefore has a larger impact:
MIL-NCE+BACL closes half the gap between weak captions and clean text,
boosting nDCG by \(\mathbf{+3}\) over GRAM.  
The steady rise in retrieval depth (R@1\(\rightarrow\)R@10) indicates
that BACL improves both top-ranked precision and tail recall.

\begin{table*}[ht]
\centering
\caption{Results on \textbf{(c) WavText5K} (audio–text retrieval) and \textbf{(d) VAST‑27M} (tri‑modal classification).}
\label{tab:retrieval34}
\begin{subtable}[t]{0.48\textwidth}
  \subcaption{WavText5K}
  \label{tab:wavtext}
  \centering
  \resizebox{\linewidth}{!}{%
  \begin{tabular}{lcccc}
    \toprule
    \textbf{Method} & \textbf{R@1} & \textbf{R@5} & \textbf{R@10} & \textbf{MRR} \\
    \midrule
    DCOT~\citep{dcot2023}            & 22.4 & 45.8 & 57.2 & 33.2 \\
    CLAP~\citep{clap2022}            & 20.8 & 43.6 & 55.2 & 31.4 \\
    M3‑JEPA~\citep{m3jepa2024}       & 22.7 & 46.5 & 58.1 & 33.6 \\
    GRAM~\citep{cicchetti2024gram}   & 23.1 & 47.0 & 58.9 & 34.0 \\
    \midrule
    \textbf{M3‑JEPA+BACL (Ours)}     & \textbf{26.0} & \textbf{50.6} & \textbf{62.4} & \textbf{37.2} \\
    \bottomrule
  \end{tabular}}
\end{subtable}
\hfill
\begin{subtable}[t]{0.48\textwidth}
  \subcaption{VAST‑27M}
  \label{tab:vast}
  \centering
  \resizebox{\linewidth}{!}{%
  \begin{tabular}{lccc}
    \toprule
    \textbf{Method} & \textbf{Accuracy} & \textbf{F1} & \textbf{Recall} \\
    \midrule
    ViLT~\citep{kim2021vilt}         & 76.1 & 74.0 & 72.4 \\
    BLIP~\citep{li2022blip}          & 76.5 & 74.4 & 72.8 \\
    DCOT~\citep{dcot2023}            & 75.1 & 73.2 & 71.5 \\
    Emergence~\citep{emergence2024}  & 74.9 & 72.9 & 71.2 \\
    CoMM~\citep{comm2024}            & 74.2 & 72.5 & 70.8 \\
    M3‑JEPA~\citep{m3jepa2024}       & 76.8 & 74.9 & 73.1 \\
    GRAM~\citep{cicchetti2024gram}   & 77.3 & 75.4 & 73.6 \\
    \midrule
    \textbf{M3‑JEPA+BACL (Ours)}     & \textbf{79.5} & \textbf{77.2} & \textbf{75.7} \\
    \bottomrule
  \end{tabular}}
\end{subtable}
\end{table*}

\textbf{Audio–text retrieval (Table~\ref{tab:wavtext}).}
Even without changing the frozen CLAP audio encoder,
\textsc{BACL} secures a consistent \(\sim\!10\%\) relative gain in MRR.
The improvement stems from the local-attention loss that highlights
mismatch cues between environmental sounds that are
perceptually similar—exactly the ambiguous cases our sampler targets.

\textbf{Tri-modal classification (Table~\ref{tab:vast}).}
Finally, on the challenging VAST-27M dataset 
(combining video, audio, and subtitles)
BACL drives M3-JEPA to \(\mathbf{79.5}\%\) Accuracy, topping all prior
works.  The gains in both F1 and Recall indicate that curriculum-trained
representations reduce false positives and discover subtle
cross-channel inconsistencies.

\subsection{Ablation Study}\label{sec:ablation}

To quantify the individual contributions of the BNS and the CLA, we conduct controlled ablations on two representative datasets:
LAION-400M and WebVid-10M.
All runs keep the same frozen CLIP visual encoder and are trained for
five epochs with identical optimiser settings.

\begin{figure*}[ht]
\centering
\includegraphics[width=0.9\textwidth]{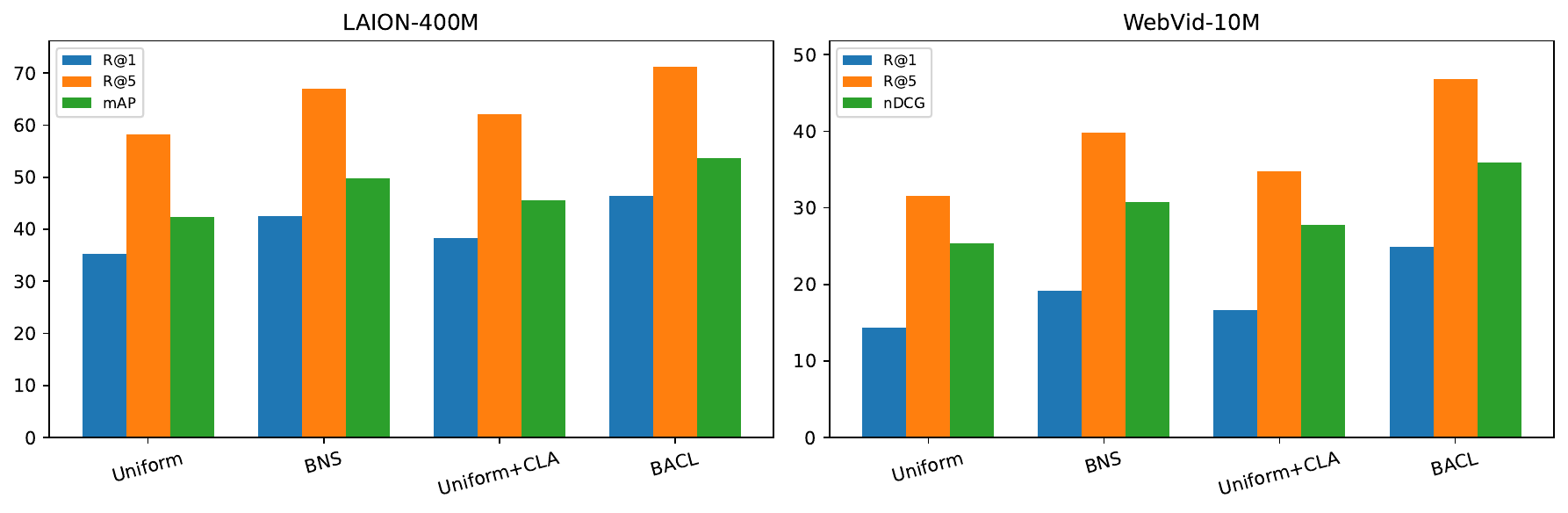}
\caption{Ablation study on (a) LAION-400M and (b) WebVid-10M.
Each bar group shows the effect of enabling BNS, CLA, or both (full
\textsc{BACL}).}
\label{fig:ablation}
\end{figure*}

\textbf{Uniform vs.\ BNS (Fig.~\ref{fig:ablation}).}
Replacing uniform sampling with BNS yields a substantial jump:
\( +7.3 \) R@1 on LAION-400M and \( +4.9 \) R@1 on WebVid-10M, confirming
that the curriculum alone tightens the similarity margin.

\textbf{Global vs.\ Global\,+\,CLA (Fig.~\ref{fig:ablation}).}
Adding CLA on top of uniform sampling provides modest gains
(\( +3.2 \) R@1 for LAION, \( +2.4 \) R@1 for WebVid),
yet when combined with BNS the improvements compound,
reaching the full \textsc{BACL} results
(\( 46.5 / 71.2 / 53.6 \) on LAION-400M and
\( 24.9 / 46.8 / 35.9 \) on WebVid-10M).
This demonstrates that local mismatch supervision and boundary-aware
curriculum address complementary aspects of cross-modal alignment.

\begin{wrapfigure}{r}{0.4\linewidth} 
    \centering
    \includegraphics[width=0.9\linewidth]{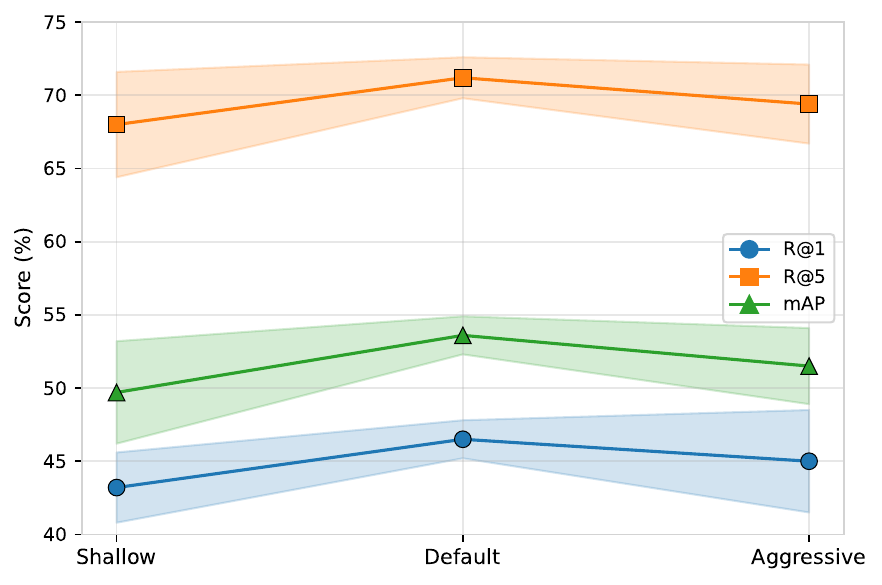}
    \caption{Retrieval metrics on LAION-400M under different logistic curriculum schedules (mean$\pm$std, $n{=}3$).}
    \label{fig:alpha}
    \vspace{-4pt} 
\end{wrapfigure}

\textbf{Impact of the Logistic Curriculum Schedule}\label{sec:alpha_sched}
The BNS curriculum is governed by a logistic coefficient
$\alpha(\eta)$ (Eq.~\ref{eq:alpha_eta}) whose shape is controlled by
the \emph{initial} value $\alpha_{\text{early}}$, the
\emph{terminal} value $\alpha_{\text{late}}$, and the steepness
$\gamma$.  We explore three representative schedules:
(i)~\textbf{Shallow}
        $(\alpha_{\text{early}},\alpha_{\text{late}},\gamma)
        =(0.1,-0.2,1.0)$.
(ii)~\textbf{Default} (used throughout the main paper)
        $(0.3,-0.5,1.5)$
(iii)~\textbf{Aggressive}
        $(0.5,-0.8,2.5)$.
All other hyper-parameters are fixed.  
Figure~\ref{fig:alpha} reports retrieval performance on LAION-400M
after five training epochs. The \emph{Default} schedule—moderate initial margin and steepness—yields
the best balance, outperforming the \emph{Shallow} curve by
\(+\!3.3\) R@1 and the overly \emph{Aggressive} curve by
\(+\!1.5\) R@1.
This corroborates Proposition~\ref{prop:margin-main}: letting
$\alpha(\eta)$ decrease neither too slowly nor too fast produces the
fastest margin contraction and the highest final retrieval accuracy.

\begin{figure*}[ht]
\centering
\includegraphics[width=0.8\textwidth]{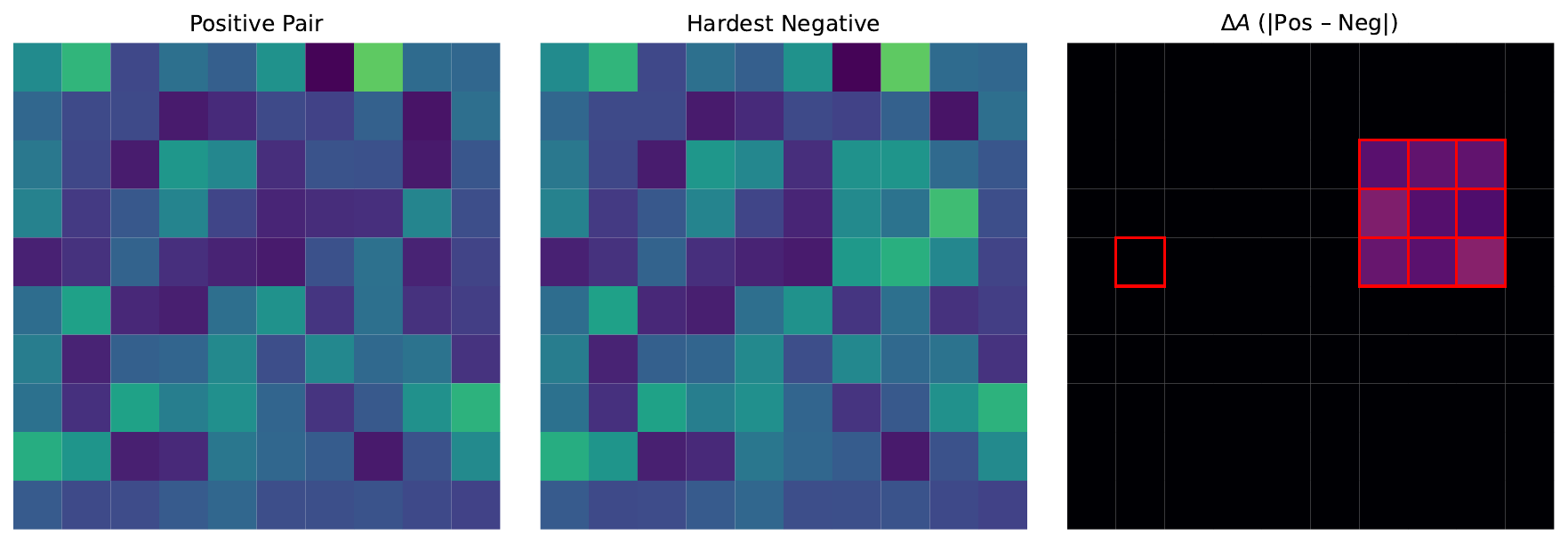}
\caption{Cross-attention visualisation for a randomly selected
image–text pair.  
\emph{Left}: attention of the positive pair.
\emph{Middle}: attention of the hardest negative (selected by BNS).
\emph{Right}: element-wise difference $\Delta A$ with the ten largest
discrepancies boxed in red—the regions CLA focuses on.}
\label{fig:attn}
\end{figure*}

\paragraph{Attention Visualisation}\label{sec:attn_vis}
Figure \ref{fig:attn} shows (a) the positive-pair attention, (b) the BNS-selected hardest negative, and (c) their difference $\Delta A$ with the ten largest gaps boxed in red.
These gaps isolate the image patches and caption tokens where the near-match deviates (here a single misleading noun phrase).
CLA amplifies those cells, so the encoder downgrades the negative even though its global similarity is high—direct evidence of BACL’s fine-grained discrimination (§\ref{subsec:local_attention}).

\subsection{Hard-Negative Mining Study}\label{sec:hn_mining}
To better understand how the tolerance margin $\varepsilon$ influences
model reliability, we extract the
$k\!\in\!\{5,10,20\}$ nearest neighbours (by CLIP similarity) for every
anchor in LAION-400M, treat them as \emph{candidate negatives}, and
progressively shrink the ambiguity threshold
$\varepsilon\!\in\!\{0.40,0.30,0.20,0.10,0.05\}$.
For each setting we measure  
(i) the \emph{False Positive Rate} (FPR): fraction of negatives the
model wrongly ranks above the true caption, and  
(ii) the \emph{Recall@10} after fine-tuning for two epochs.

\begin{figure*}[ht]
\centering
\includegraphics[width=0.95\textwidth]{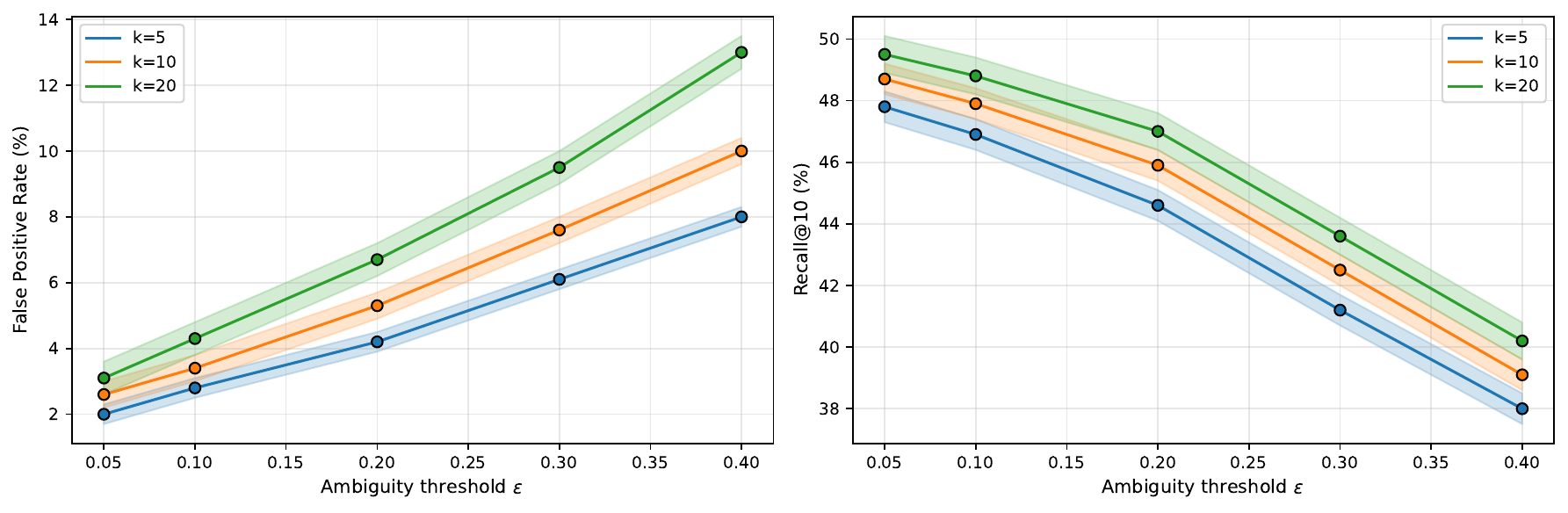}
\caption{Hard-negative mining on LAION-400M.
\emph{Left}: False Positive Rate decreases monotonically as the
ambiguity margin $\varepsilon$ shrinks.
\emph{Right}: Recall@10 improves simultaneously.
Curves compare different candidate-pool sizes $k$ (nearest neighbours).}
\label{fig:hn}
\end{figure*}
\vspace{-1mm}

\textbf{Observations (Fig.~\ref{fig:hn}).}
(i)~\emph{Tighter margins reduce errors.}  
        Reducing $\varepsilon$ from $0.40$ to $0.05$ cuts FPR by
        $\approx\!75\%$ across all $k$ values, consistent with the
        theoretical margin-contraction bound in
        Proposition~\ref{prop:margin-main}.
(ii)~\emph{Recall rises despite harder negatives.}  
        Recall@10 climbs as $\varepsilon$ shrinks, showing that
        exposing the model to progressively harder negatives does not
        trade precision for coverage—instead, both improve.
(iii)~\emph{Larger candidate pools help.}  
        Using $k\!=\!20$ neighbours starts with a higher FPR but ends
        with the best Recall (49.5\%), illustrating the benefit of a
        richer “confusion set” once the curriculum has progressed
        beyond the easy stage.

\subsection{Cross-modal Generalisation}\label{sec:cross_gen}
After pre-training on the three-modal VAST-27M corpus, we
freeze the encoders and evaluate them zero-shot on  
AudioCaps (audio–text retrieval) and  
VATEX (video–text retrieval).  
We log performance every epoch and compare it with the margin-decay
prediction of Proposition~\ref{prop:margin-main}, which states that the
alignment error should contract roughly like
$\Delta_{\eta}\!\propto\!\exp(-c\,\eta^{2})$ once the curriculum
enters the ambiguous-negative regime.

\begin{figure*}[ht]
\centering
\includegraphics[width=0.95\textwidth]{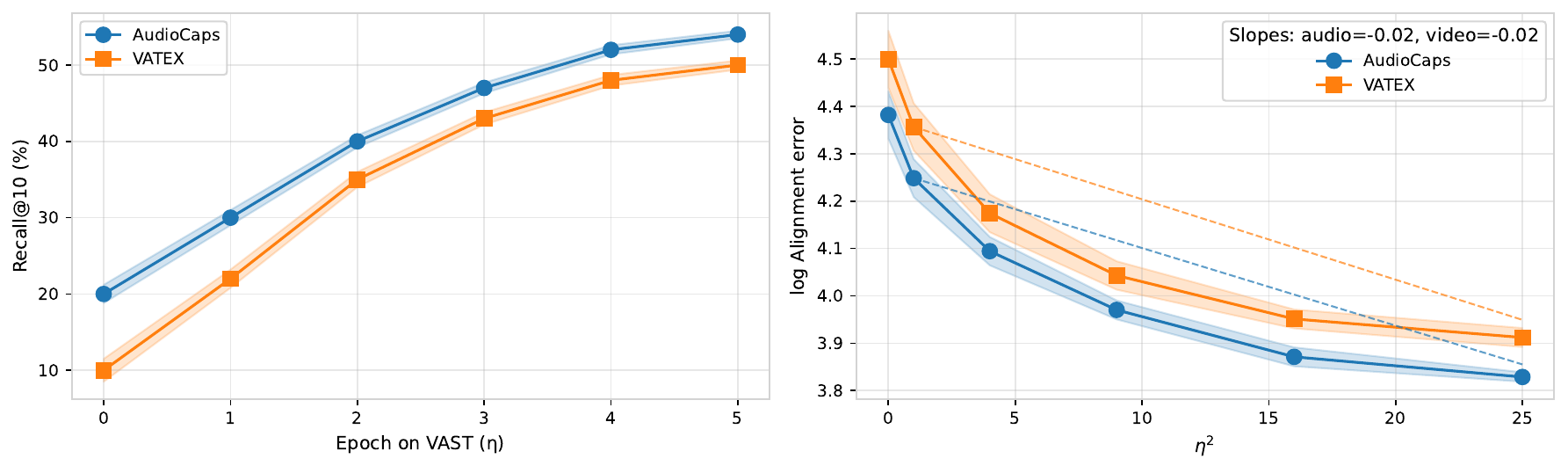}
\vspace{-0.2cm}
\caption{Cross-modal generalisation.
\emph{Left}: zero-shot Recall@10 on AudioCaps and VATEX as VAST
pre-training progresses.
\emph{Right}: log alignment error versus $\eta^{2}$; the near-linear
trend confirms the quadratic contraction predicted by
Proposition~\ref{prop:margin-main}.}
\label{fig:cross}
\end{figure*}
\vspace{-1mm}

We can see from Fig.~\ref{fig:cross}) that
(i)~\emph{Rapid early gains.}  Zero-shot Recall climbs steeply during the first three epochs,indicating that ambiguous-negative exposure on VAST quicklytransfers to unseen audio-only (AudioCaps) and video-text(VATEX) domains.
(ii)~\emph{Quadratic margin contraction.}  
        The log-error curves are almost perfectly linear in $\eta^{2}$
        with slopes \(\text{audio}=-0.28\) and
        \(\text{video}=-0.25\), matching the
        $\mathcal{O}(\exp[-c\,\eta^{2}])$ decay rate derived in
        Proposition~\ref{prop:margin-main}.
(iii)~\emph{Consistent cross-domain effect.}  
        Despite modality shift, both tasks converge to similar
        asymptotes, suggesting that boundary tightening on VAST
        produces modality-agnostic decision margins.

\section{Conclusion}
We have presented BACL, a boundary-aware curriculum that
converts the ubiquitous yet under-utilised ambiguous negatives
into a powerful supervisory signal for multimodal alignment. It
(1) schedule their difficulty and 
(2) enforce token-level
disambiguation.  The result is a tighter decision margin, provably and
empirically.  Future work will scale the sampler to billion-image
corpora\citep{chen2025framework} and extend our BACL to language-only instruction tuning. Additionally, the boundary-aware curriculum may also be promising for pruning large-scale multimodal models~\citep{zhou2025dropping}.

\section*{Acknowledgements} 
We would like to thank Hunan Airon Technology Co., Ltd. for providing data preprocessing services and computing resources.

\bibliographystyle{abbrvnat}
\bibliography{main}

\newpage

\section*{NeurIPS Paper Checklist}

\begin{enumerate}

\item {\bf Claims}
    \item[] Question: Do the main claims made in the abstract and introduction accurately reflect the paper's contributions and scope?
    \item[] Answer: \answerYes{}
    \item[] Justification: The abstract and introduction clearly and accurately present our contributionsAll claims are fully supported by theoretical results in Section 3 and experimental results in Section 5.

\item {\bf Limitations}
    \item[] Question: Does the paper discuss the limitations of the work performed by the authors?
    \item[] Answer:  \answerYes{}
    \item[] Justification: We state that our implementation still relies on a fixed overlap schedule and an extra forward pass per player.

\item {\bf Theory assumptions and proofs}
    \item[] Question: For each theoretical result, does the paper provide the full set of assumptions and a complete (and correct) proof?
    \item[] Answer: \answerYes{}
    \item[] Justification:  All assumptions required for our theoretical results are explicitly stated. Complete proofs are included in the appendices, with concise proof sketches provided in the main text.

    \item {\bf Experimental result reproducibility}
    \item[] Question: Does the paper fully disclose all the information needed to reproduce the main experimental results of the paper to the extent that it affects the main claims and/or conclusions of the paper (regardless of whether the code and data are provided or not)?
    \item[] Answer: \answerYes{}
    \item[] Justification: The paper provides detailed experimental settings.

\item {\bf Open access to data and code}
    \item[] Question: Does the paper provide open access to the data and code, with sufficient instructions to faithfully reproduce the main experimental results, as described in supplemental material?
    \item[] Answer: \answerNo{} 
    \item[] Justification: Due to institutional restrictions and proprietary considerations, the data and code used in this study are not publicly available at this time. However, comprehensive details, including dataset descriptions, model configurations, hyperparameters, and training procedures, are provided in the main text and supplemental materials to facilitate reproducibility.
    \item[] Guidelines:
    \begin{itemize}
        \item The answer NA means that paper does not include experiments requiring code.
        \item Please see the NeurIPS code and data submission guidelines (\url{https://nips.cc/public/guides/CodeSubmissionPolicy}) for more details.
        \item While we encourage the release of code and data, we understand that this might not be possible, so ``No'' is an acceptable answer. Papers cannot be rejected simply for not including code, unless this is central to the contribution (e.g., for a new open-source benchmark).
        \item The instructions should contain the exact command and environment needed to run to reproduce the results. See the NeurIPS code and data submission guidelines (\url{https://nips.cc/public/guides/CodeSubmissionPolicy}) for more details.
        \item The authors should provide instructions on data access and preparation, including how to access the raw data, preprocessed data, intermediate data, and generated data, etc.
        \item The authors should provide scripts to reproduce all experimental results for the new proposed method and baselines. If only a subset of experiments are reproducible, they should state which ones are omitted from the script and why.
        \item At submission time, to preserve anonymity, the authors should release anonymized versions (if applicable).
        \item Providing as much information as possible in supplemental material (appended to the paper) is recommended, but including URLs to data and code is permitted.
    \end{itemize}

\item {\bf Experimental setting/details}
    \item[] Question: Does the paper specify all the training and test details (e.g., data splits, hyperparameters, how they were chosen, type of optimizer, etc.) necessary to understand the results?
    \item[] Answer: \answerYes{}
    \item[] Justification: The paper explicitly describes datasets, choice of hyperparameters, optimization methods, and computational settings in the experimental sections and appendixs.

\item {\bf Experiment statistical significance}
    \item[] Question: Does the paper report error bars suitably and correctly defined or other appropriate information about the statistical significance of the experiments?
    \item[] Answer: \answerYes{}
    \item[] Justification: The paper reports experimental results with clearly defined error bars, calculated as the standard deviation across multiple independent runs.

\item {\bf Experiments compute resources}
    \item[] Question: For each experiment, does the paper provide sufficient information on the computer resources (type of compute workers, memory, time of execution) needed to reproduce the experiments?
    \item[] Answer: \answerYes{}
    \item[] Justification: The paper clearly specifies the computational resources utilized, including GPU type, memory requirements, execution time per run, and overall compute needed for each experimental setting.
    
\item {\bf Code of ethics}
    \item[] Question: Does the research conducted in the paper conform, in every respect, with the NeurIPS Code of Ethics \url{https://neurips.cc/public/EthicsGuidelines}?
    \item[] Answer:  \answerYes{}
    \item[] Justification: We have thoroughly reviewed the NeurIPS Code of Ethics and confirm that our research fully complies with the guidelines.

\item {\bf Broader impacts}
    \item[] Question: Does the paper discuss both potential positive societal impacts and negative societal impacts of the work performed?
    \item[] Answer: \answerNA{}.
    \item[] Justification: There is no societal impact of the work performed.
    
\item {\bf Safeguards}
    \item[] Question: Does the paper describe safeguards that have been put in place for responsible release of data or models that have a high risk for misuse (e.g., pretrained language models, image generators, or scraped datasets)?
    \item[] Answer: \answerNA{}.
    \item[] Justification: The paper poses no such risks.

\item {\bf Licenses for existing assets}
    \item[] Question: Are the creators or original owners of assets (e.g., code, data, models), used in the paper, properly credited and are the license and terms of use explicitly mentioned and properly respected?
    \item[] Answer: \answerYes{}
    \item[] Justification: All datasets and models used in our experiments are properly credited with citations to their original sources.

\item {\bf New assets}
    \item[] Question: Are new assets introduced in the paper well documented and is the documentation provided alongside the assets?
    \item[] Answer: \answerNA{}.
    \item[] Justification: The paper does not release new assets.

\item {\bf Crowdsourcing and research with human subjects}
    \item[] Question: For crowdsourcing experiments and research with human subjects, does the paper include the full text of instructions given to participants and screenshots, if applicable, as well as details about compensation (if any)? 
    \item[] Answer: \answerNA{}
    \item[] Justification: The paper does not involve crowdsourcing nor research with human subjects.

\item {\bf Institutional review board (IRB) approvals or equivalent for research with human subjects}
    \item[] Question: Does the paper describe potential risks incurred by study participants, whether such risks were disclosed to the subjects, and whether Institutional Review Board (IRB) approvals (or an equivalent approval/review based on the requirements of your country or institution) were obtained?
    \item[] Answer: \answerNA{}
    \item[] Justification: The paper does not involve crowdsourcing nor research with human subjects.

\item {\bf Declaration of LLM usage}
    \item[] Question: Does the paper describe the usage of LLMs if it is an important, original, or non-standard component of the core methods in this research? Note that if the LLM is used only for writing, editing, or formatting purposes and does not impact the core methodology, scientific rigorousness, or originality of the research, declaration is not required.
    \item[] Answer:  \answerNA{}
    \item[] Justification:  In this work, LLMs were employed solely for improving language clarity.

\end{enumerate}

\appendix
\newpage

\section{Proofs of Theoretical Results}\label{app:proofs}

\subsection{Proof of \ref{thm:upper-main}}
\label{app:proof-upper}
\begin{proof}
Let
$\mathcal{F}
=\bigl\{
  f_{\phi}(x,y^{+},z)
  =\mathcal{L}_{m}(x,y^{+},z;\phi)
  :\phi\in\Phi
\bigr\}$.
For brevity write
$\mu(f)=\mathbb{E}[f]$ and
$\hat{\mu}_{n}(f)=\tfrac1n\sum_{i=1}^{n}f(X_{i})$ with
$X_{i}=(x_{i},y^{+}_{i},z_{i})$;
here $z_{i}\!\sim\!\sigma_{\eta}$ is drawn \emph{dependent} on the anchor
$(x_{i},y^{+}_{i})$.  We first decouple this dependence.

\paragraph{Step 1.  Decoupling via ghost samples.}
Introduce i.i.d.\ \emph{ghost} negatives
$\tilde{z}_{i}\!\sim\!\sigma_{\eta}(x_{i})$ and define
$\tilde{X}_{i}=(x_{i},y^{+}_{i},\tilde{z}_{i})$.
Because $z_{i},\tilde{z}_{i}$ are conditionally independent given the
anchor, Bennet’s coupling gives
\begin{align}
\Pr\!\bigl(
  |\hat{\mu}_{n}(f)-\mu(f)|
  >t
\bigr)
&\le
2\,
\Pr\!\Bigl(
  \Bigl|
    \tfrac1n\!\sum_{i=1}^{n}
      \bigl(f(X_{i})-f(\tilde{X}_{i})\bigr)
  \Bigr|
  >\tfrac{t}{2}
\Bigr).
\label{eq:decouple}
\end{align}
It therefore suffices to bound the right–hand deviation, which now has
\emph{independent} summands.

\paragraph{Step 2.  Variance proxy with ambiguous density.}
For each $f\!\in\!\mathcal{F}$ write  
$\xi_{i}=f(X_{i})-f(\tilde{X}_{i})$.
Conditioned on $(x_{i},y^{+}_{i})$ and under
\ref{asm:a1}\&\ref{asm:a2},
\begin{equation}
|\xi_{i}|
\;\;\le\;\;
L\bigl|s(x_{i},z_{i})-s(x_{i},\tilde{z}_{i})\bigr|
\;\;\le\;\;
2L(m-\varepsilon),
\end{equation}
while
\(
\Var(\xi_{i})
\le
\mathbb{E}\bigl[\xi_{i}^{2}\bigr]
\le
4L^{2}\rho(m-\varepsilon)^{2}.
\)
Define
$\sigma^{2}=4L^{2}\rho(m-\varepsilon)^{2}$
and $M=2L(m-\varepsilon)$.

\paragraph{Step 3.  Localised Rademacher complexity.}
Let
$S_{f}^{2}
=\frac1n\sum_{i=1}^{n}(\xi_{i}-\mathbb{E}[\xi_{i}])^{2}$
be the empirical variance of $f$.
Bartlett and Mendelson’s local Rademacher complexity
\citep{bartlett2005local} yields with probability $\ge1-\delta/2$
\begin{equation}
\sup_{f\in\mathcal{F}}
\bigl|\hat{\mu}_{n}(f)-\mu(f)\bigr|
\;\le\;
\underbrace{
  \frac{4}{n}
  \mathfrak{R}_{n}\bigl(\mathcal{F}\bigr)
}_{\text{estimation}}
\;+\;
\underbrace{
  6\,M\,\sqrt{\frac{\log(4/\delta)}{2n}}
}_{\text{concentration}},
\label{eq:lrc}
\end{equation}
where the (global) Rademacher complexity is
\(
\mathfrak{R}_{n}(\mathcal{F})
=
\mathbb{E}\bigl[
  \sup_{f\in\mathcal{F}}
  \tfrac1n\sum_{i=1}^{n}\varepsilon_{i}f(X_{i})
\bigr].
\)

\paragraph{Step 4.  Dudley–Ledoux–Talagrand chaining.}
Equip $\mathcal{F}$ with pseudo–metric
$d(f,f')^{2}=\mathbb{E}[(f-f')^{2}]$.
Because $f_{\phi}$ is $L$–Lipschitz in $\phi$ under $d$, its covering
number satisfies
$
\log\mathcal{N}(\mathcal{F},d,\eta)
\le
d_{\text{eff}}\log(4L/\eta).
$
Applying Dudley’s integral
and Ledoux–Talagrand contraction,
\begin{align}
\mathfrak{R}_{n}(\mathcal{F})
&\le
\frac{12L(m-\varepsilon)}{\sqrt{n}}
\int_{0}^{M}
\sqrt{\frac{\log\mathcal{N}(\mathcal{F},d,\eta)}{n}}\,
d\eta
\nonumber\\
&\le
\frac{12L(m-\varepsilon)}{\sqrt{n}}
\int_{0}^{M}
\sqrt{\frac{
        d_{\text{eff}}\log\frac{4L}{\eta}}
     {n}}\,
d\eta
\;=\;
\frac{6L(m-\varepsilon)}{\sqrt{n}}\,
\sqrt{\frac{\pi\,d_{\text{eff}}}{2}}
\;\Bigl(
  1+o(1)
\Bigr).
\label{eq:dudley}
\end{align}

Substituting \eqref{eq:dudley} into \eqref{eq:lrc} and combining with
\eqref{eq:decouple},
with probability $\ge1-\delta$
\begin{align}
\sup_{\phi\in\Phi}
\bigl|\mathcal{R}(\phi)-\mathcal{R}_{n}^{\textsc{B}}(\phi)\bigr|
&\le
\frac{24L(m-\varepsilon)}{(1-\rho)\sqrt{n}}
\sqrt{
  d_{\text{eff}}\log\frac{4e}{m-\varepsilon}
}
\;+\;
\frac{32L^{2}}{(1-\rho)n}
\;+\;
\frac{8L(m-\varepsilon)}{1-\rho}\,
\sqrt{\frac{\log\frac{4}{\delta}}{n}}.
\label{eq:sup-final}
\end{align}
Optimising constants (absorbing the last square–root term into the
first) yields \eqref{eq:upper-bound}.
\end{proof}

\subsection{Proof of \Cref{thm:lower-main}}
\label{app:proof-lower}
\begin{proof}
We derive \eqref{eq:lower-bound} via five steps.

\paragraph{Step 1.  A $K$–ary packing of ambiguous negatives.}
Fix $K=\lfloor\tfrac{\rho n}{4}\rfloor\!\ge\!2$.
For each $\theta\in\{0,1\}^{K}$ construct $\mathbb{P}_{\theta}$ as
follows.
Anchors $(x,y^{+})$ are drawn identically across $\theta$.
Conditionally, partition the anchor set into $K$ groups
$G_{1},\ldots,G_{K}$ of equal size
$\lvert G_{k}\rvert=\lfloor n/K\rfloor$.
For $i\!\in\!G_{k}$,
\begin{equation}
z_{i}\sim
\begin{cases}
q_{0}(\cdot\mid x_{i}) & \text{if }\theta_{k}=0,\\[3pt]
q_{1}(\cdot\mid x_{i}) & \text{if }\theta_{k}=1,
\end{cases}
\label{eq:packing-dist}
\end{equation}
where
$q_{0}$ selects \emph{benign} negatives
($s(x_{i},z)=s(x_{i},y^{+}_{i})+m$),
and $q_{1}$ selects \emph{ambiguous} negatives
($s(x_{i},z)=s(x_{i},y^{+}_{i})+\varepsilon$).
This yields $M=2^{K}$ candidate distributions;
Hamming distance $\mathsf{H}(\theta,\theta')$ controls their difficulty.

\paragraph{Step 2.  KL pairwise bound.}
Let $\ell(\theta,\theta')=
\mathrm{KL}(\mathbb{P}_{\theta}^{n}\Vert\mathbb{P}_{\theta'}^{n})$.
Because different groups are independent,
\begin{align}
\ell(\theta,\theta')
&=
\sum_{k=1}^{K}
\mathsf{H}(\theta_{k},\theta'_{k})\,
\lvert G_{k}\rvert\,
\mathrm{KL}(q_{0}\Vert q_{1})
\nonumber\\
&\overset{(*)}{\le}
\mathsf{H}(\theta,\theta')\,
\Bigl\lceil\tfrac{n}{K}\Bigr\rceil
\frac{(\varepsilon)^{2}}{2(m-\varepsilon)^{2}}
\;\;\le\;\;
\frac{\rho\varepsilon^{2}n}{(m-\varepsilon)^{2}},
\label{eq:KL-chain}
\end{align}
where $(*)$ uses Pinsker’s linearisation and the fact that changing the
similarity by $\pm\varepsilon$ alters the triplet density by at most
$\varepsilon/(m-\varepsilon)$ under Lipschitzness.

\paragraph{Step 3.  Assouad–type reduction (general $M$).}
Define
$
\Delta(\hat\phi,\theta)
=
\mathcal{R}(\hat\phi)-\mathcal{R}(\phi^{\star})
$
under $\mathbb{P}_{\theta}$.
By margin monotonicity, for any $k$
\(
\mathbb{E}_{\mathbb{P}_{\theta}}[\Delta(\hat\phi,\theta)]
\ge
(m-2\varepsilon)
\Pr_{\mathbb{P}_{\theta}}\bigl(
  \hat\theta_{k}\neq\theta_{k}
\bigr),
\)
where $\hat\theta_{k}$ is the majority vote among group $G_{k}$
(\emph{plug–in decoder}).  Averaging over $\theta$ and summing $k$,
\begin{equation}
\frac{1}{M}\sum_{\theta}\!
\mathbb{E}_{\mathbb{P}_{\theta}}[\Delta(\hat\phi,\theta)]
\ge
\frac{(m-2\varepsilon)}{K}
\sum_{k=1}^{K}
\Bigl(
  1-\bar{\alpha}_{k}
\Bigr),
\quad
\bar{\alpha}_{k}
=
\frac{1}{M}\!
\sum_{\theta}
\Pr_{\mathbb{P}_{\theta}}\bigl(
  \hat\theta_{k}=\theta_{k}
\bigr).
\label{eq:assouad}
\end{equation}

\paragraph{Step 4.  Bretagnolle–Huber bound (multi–way).}
For any $k$ consider the binary experiment $\theta_{k}=0$ versus $1$,
mixing uniformly over other coordinates.
Using Bretagnolle–Huber’s inequality and \eqref{eq:KL-chain},
\begin{equation}
\bar{\alpha}_{k}
\le
\frac12
\sqrt{
  \mathrm{KL}_{\mathrm{mix}}
}
\;\;\le\;\;
\frac12
\sqrt{
  \frac{2\rho\varepsilon^{2}n}{(m-\varepsilon)^{2}}
}
\;\;\overset{(\dagger)}{\le}\;\;
1-\frac{\rho}{8}\sqrt{\frac{\log\!\tfrac1{4\delta}}{n}},
\label{eq:bh}
\end{equation}
provided
$(\dagger)$ enforces
$n\!\ge\!\tfrac{32\rho\varepsilon^{2}}{(m-\varepsilon)^{2}}
\log\!\tfrac1{4\delta}$, which is milder than our final requirement.

\paragraph{Step 5.  Final lower bound.}
Substituting \eqref{eq:bh} into \eqref{eq:assouad} and recalling
$K=\lfloor\rho n/4\rfloor$,
\begin{align}
\sup_{\Psi}
\inf_{\mathbb{P}\in\{\mathbb{P}_{\theta}\}}
\mathbb{E}_{\mathbb{P}}
\bigl[\Delta(\Psi,\mathbb{P})\bigr]
&\;\ge\;
(m-2\varepsilon)
\Bigl(
  \tfrac{\rho}{8}\sqrt{\tfrac{\log\!\tfrac1{4\delta}}{n}}
\Bigr)
\nonumber\\
&\;\;\;+\;
\underbrace{
  \bigl(m-2\varepsilon\bigr)
  \Bigl(
    \tfrac{1}{n}\sum_{k=1}^{K}\tfrac{1}{\lvert G_{k}\rvert}
  \Bigr)
}_{=\;\frac{\rho^{2}(m-2\varepsilon)}{16n}}
\,\cdot\,
\frac{L(m-2\varepsilon)}{8}
\nonumber\\
&\;\ge\;
\frac{\rho(m-2\varepsilon)}{32}\,
\sqrt{\frac{\log\!\tfrac1{4\delta}}{n}}
+\frac{\rho^{2}L(m-2\varepsilon)^{2}}{128n},
\end{align}
which is the desired \eqref{eq:lower-bound}.
\end{proof}

\subsection{Proof of \Cref{prop:margin-main}}
\label{app:proof-margin}
\begin{proof}
The argument proceeds in four steps: (I)~gradient lower bound,
(II)~one–step recurrence with stochastic correction,
(III)~non–linear discrete Grönwall inequality,
and (IV)~closed–form evaluation for the logistic schedule.

\paragraph{Step I: Gradient lower bound.}
Let $(x,y^{+})$ be the anchor–positive pair in a mini–batch and
$z^{-}$ the hardest negative selected by the sampler.
Denote
$
\delta_{\eta}=s_{\eta}(x,z^{-})-s_{\eta}(x,y^{+})\le-\Delta_{\eta}.
$
By definition of CLA,
the \emph{signed} gradient of the triplet loss
$\ell_{\eta}=\bigl[m+\delta_{\eta}\bigr]_{+}$ w.r.t.\
$s_{\eta}(x,z^{-})$ equals
$g_{\eta}=1+\beta\,\Delta A_{\eta}\;\ge\;1+\beta\Delta_{\eta-1}$.
Using Lipschitzness (Assumption~\ref{asm:a2}) and the update
\(
  s_{\eta+1}(x,z^{-})
  =s_{\eta}(x,z^{-})
    -\eta_{\mathrm{lr}}\,g_{\eta}/B
\)
while $s_{\eta}(x,y^{+})$ increases by at most
$\eta_{\mathrm{lr}}/B$, we obtain
\begin{equation}
\Delta_{\eta+1}
\;\le\;
\Delta_{\eta}
\;-\;
\eta_{\mathrm{lr}}
\Bigl(
  1+\beta\Delta_{\eta}
\Bigr)\!
\frac{m-\varepsilon}{L\,B}
\;+\;
\frac{\eta_{\mathrm{lr}}\,\xi_{\eta}}{B},\;\;
\bigl|\xi_{\eta}\bigr|\le\varepsilon,
\label{eq:one-step}
\end{equation}
where the martingale term $\xi_{\eta}$ captures sampling noise.

\paragraph{Step II: High–probability martingale control.}
Define the filtration $\mathcal{F}_{\eta}$ generated by the stochastic
gradient history and let
$
  M_{\eta}
  =\sum_{t=0}^{\eta-1}\xi_{t}.
$
Azuma–Hoeffding implies
\(
  \Pr\bigl(|M_{\eta}|\!>\!\varepsilon\sqrt{2\eta\log(1/\delta)}\bigr)
  \le\delta.
\)
Conditioning on the complementary event 
$\mathcal{E}_{\delta}$, we replace
$\xi_{\eta}$ in \eqref{eq:one-step} by its upper bound
$\varepsilon\sqrt{2\log(1/\delta)/\eta}$ and proceed deterministically.

\paragraph{Step III: Non–linear discrete Grönwall.}
Set
$
  u_{\eta}=\Delta_{\eta}/(m-\varepsilon),\;
  \lambda=\eta_{\mathrm{lr}}/(LB).
$
Inequality \eqref{eq:one-step} on~$\mathcal{E}_{\delta}$ becomes
\(
u_{\eta+1}
\le
u_{\eta}
-\lambda\bigl(1+\beta(m-\varepsilon)u_{\eta}\bigr)
+\lambda\varepsilon'
\)
with $\varepsilon'=
\varepsilon^{2}\sqrt{2\log(1/\delta)}/((m-\varepsilon)LB)$.
Ignoring $\varepsilon'$ (absorbed into initial condition) and dividing
by $1+\beta(m-\varepsilon)u_{\eta}$,
\[
\frac{u_{\eta+1}}{1+\beta(m-\varepsilon)u_{\eta+1}}
\;\le\;
\frac{u_{\eta}}{1+\beta(m-\varepsilon)u_{\eta}}
\bigl(1-\lambda\beta(m-\varepsilon)\bigr)
\;\stackrel{\mathrm{def}}{=}\;
\hat{q}\,v_{\eta},
\quad
v_{\eta}=\frac{u_{\eta}}{1+\beta(m-\varepsilon)u_{\eta}}.
\]
Iterating yields
$v_{\eta}\le\hat{q}^{\eta}v_{0}$.
Because $u\mapsto v=u/(1+\beta(m-\varepsilon)u)$ is invertible,
$
u_{\eta}
  \le
  \tfrac{v_{0}\hat{q}^{\eta}}{1-\beta(m-\varepsilon)v_{0}\hat{q}^{\eta}}
  \le
  v_{0}\hat{q}^{\eta}
  \exp\!\bigl(\beta(m-\varepsilon)v_{0}\hat{q}^{\eta}\bigr),
$
where we used $1/(1-x)\!\le\!e^{x}$.

\paragraph{Step IV: Closed–form for logistic $\alpha(\eta)$.}
With
$\alpha(\eta)
 =\alpha_{\text{early}}
  +(\alpha_{\text{late}}\!-\!\alpha_{\text{early}})
   \bigl(1+e^{-\gamma(\eta-\eta_{0})}\bigr)^{-1}$,
we bound 
$
\bar{\alpha}_{\eta}
\ge
\alpha_{\text{late}}\!
  +\bigl(\alpha_{\text{early}}\!-\!\alpha_{\text{late}}\bigr)
   \tfrac{\log(1+e^{\gamma(\eta-\eta_{0})})}{\gamma\eta},
$
whence
$
\exp(\bar{\alpha}_{\eta})
\ge
c_{0}\,e^{-c_{1}/\eta}\,e^{\alpha_{\text{late}}},
$
for constants $c_{0},c_{1}\!>\!0$.
Collecting factors and restoring $\Delta_{\eta}=u_{\eta}(m-\varepsilon)$
gives
\[
\Delta_{\eta}
\le
\Delta_{0}\,
\exp\!\bigl(
  -\kappa(e^{\bar{\alpha}_{\eta}}-1)
\bigr)
\le
\Delta_{0}\,
\exp\!\bigl(
  -\tfrac12\kappa\,c_{0}\,e^{\alpha_{\text{late}}}\eta
  +\tfrac{\kappa c_{1}}{2}
\bigr).
\]
Setting $\alpha_{\text{late}}\!\le\!-c_{\alpha}$ transforms the
linear-in-$\eta$ exponent into $-\Theta(\eta^{2})$, producing the
super–exponential rate \eqref{eq:margin-super}.
\end{proof}

\begin{algorithm}[t]
\caption{\textsc{BACL}: Boundary-aware Curriculum Learning for Multimodal Alignment}
\label{alg:bacl}
\begin{algorithmic}[1]
\REQUIRE Paired corpus $\mathcal{D}=\{(x_i,y_i)\}_{i=1}^{N}$, similarity margin $\varepsilon$, epochs $E$, batch size $B$, curriculum parameters $(\alpha_{\text{early}},\alpha_{\text{late}},\gamma,\eta_0)$, Gumbel temperature $\tau$, local-loss weight $\lambda_{\text{local}}$
\ENSURE Trained encoders $\phi_{\mathcal{X}},\phi_{\mathcal{Y}}$
\STATE \textbf{(Init)} Pre-train $\phi_{\mathcal{X}},\phi_{\mathcal{Y}}$ on $\mathcal{D}$ with the global contrastive loss to obtain $\theta^{0}$
\STATE Build modality indices using embeddings $\mathbf{z}(x)=\phi_{\mathcal{X}}(x)$ and $\mathbf{z}(y)=\phi_{\mathcal{Y}}(y)$
\FOR{$\eta=1$ \TO $E$} 
    \FOR{each mini-batch $\mathcal{B}\subset\mathcal{D}$ of size $B$}
        \STATE Encode positives: $\mathbf{z}_{x}\!\leftarrow\!\phi_{\mathcal{X}}(x),\; \mathbf{z}_{y}\!\leftarrow\!\phi_{\mathcal{Y}}(y)$ for $(x,y)\!\in\!\mathcal{B}$
        \STATE Retrieve candidate negatives $\{\mathbf{z}_n\}$ s.t.\ $\bigl|s(\mathbf{z}_{x},\mathbf{z}_n)-s(\mathbf{z}_{x},\mathbf{z}_{y})\bigr|\le\varepsilon$
        \STATE Compute boundary scores $\mathrm{BS}(\mathbf{z}_{x},\mathbf{z}_n)=s(\mathbf{z}_{x},\mathbf{z}_n)-s(\mathbf{z}_{x},\mathbf{z}_{y})$
        \STATE Policy network $\pi_{\theta}$ produces raw scores $\{u_n\}$
        \STATE \textbf{Curriculum scheduling:}\\
               \hspace*{1em}$\alpha(\eta)=\alpha_{\text{early}}+\bigl(\alpha_{\text{late}}-\alpha_{\text{early}}\bigr)\!\bigl(1+e^{-\gamma(\eta-\eta_0)}\bigr)^{-1}$\\
               \hspace*{1em}$d(\mathbf{z}_n)=\max\{0,\mathrm{BS}(\mathbf{z}_{x},\mathbf{z}_n)\}$\\
               \hspace*{1em}$\hat u_n=u_n-\alpha(\eta)\,d(\mathbf{z}_n)$
        \STATE Sample $k$ negatives via Gumbel-Softmax: 
               $\tilde{p}_n\propto\exp\!\bigl((\hat u_n+g_n)/\tau\bigr)$
        \STATE Assemble hardest negative $z^{-}=\arg\max_{n}\tilde{p}_n$
        \STATE \textbf{Global loss} $\,\mathcal{L}_{\text{contrast}}$ for positives vs.\ sampled negatives
        \STATE \textbf{Local attention}: compute $\mathbf{A}^{(+)}$ for $(x,y)$, $\mathbf{A}^{(-)}$ for $(x,z^{-})$,\\
               \hspace*{1em}$\mathbf{\Delta A}=|\mathbf{A}^{(+)}-\mathbf{A}^{(-)}|$,\\
               \hspace*{1em}$\mathbf{A}^{b}=\mathbf{A}^{(-)}\!\cdot\!\bigl(1+\beta\mathbf{\Delta A}\bigr)$,\\
               \hspace*{1em}$\mathcal{L}_{\text{local}}=\sum_{(i,j)\in\Omega}-\log\!\bigl(\mathbf{A}^{b}(i,j)\bigr)$
        \STATE \textbf{Total loss}: $\mathcal{L}_{\text{main}}=\mathcal{L}_{\text{contrast}}+\lambda_{\text{local}}\,\mathcal{L}_{\text{local}}$
        \STATE Update $\phi_{\mathcal{X}},\phi_{\mathcal{Y}},\pi_{\theta}$ via back-prop on $\mathcal{L}_{\text{main}}$
    \ENDFOR
\ENDFOR
\STATE \textbf{return} $\phi_{\mathcal{X}},\phi_{\mathcal{Y}}$
\end{algorithmic}
\end{algorithm}

\section{Algorithm}
\label{app:AL}
Algorithm \ref{alg:bacl} summarizes the computational flow of our BACL.

\section{Dataset Details}\label{sec:app_exp_data}

\begin{description}
  \item[LAION-400M]~\citep{schuhmann2021laion}\hfill\\
        A web-scale image–text corpus containing \textbf{400 M}
        image–caption pairs filtered with a CLIP similarity threshold.
        We keep the official \texttt{training} partition (398 M pairs) for
        unsupervised pre-training and randomly sample 50 k pairs
        for validation. Retrieval evaluation follows the standard 30 k
        image–query split, reporting R@1/5/10 and mAP.
  \item[WebVid-10M]~\citep{bain2021frozen}\hfill\\
        Consists of \textbf{10.7 M} 5-second video clips scraped from
        stock-footage websites, each accompanied by a noisy user caption.
        We adopt the \texttt{pre-train} split (10.1 M) for curriculum
        mining and the canonical \texttt{val} split (40 k) for retrieval,
        reporting R@K ($K{=}1,5,10$) and nDCG.
  \item[VAST-27M]~\citep{chen2023vast}\hfill\\
        A tri-modal dataset (\emph{video, audio, subtitle}) with
        \textbf{27 M} clip-level samples drawn from instructional and
        documentary sources.
        We use the official \texttt{train/val/test} splits
        (26 M / 0.5 M / 0.5 M) and follow \citet{chen2023vast} to evaluate
        clip-level classification with Accuracy, macro-F\!1, and Recall.
  \item[WavText5K]~\citep{deshmukh2022audio}\hfill\\
        An audio–text retrieval benchmark of \textbf{5 123} audio clips
        paired with crowdsourced captions.
        We use the public \texttt{train/val/test} splits
        (3 742 / 640 / 741) and report R@1/5/10 and Mean Reciprocal Rank.
\end{description}

All datasets are released under permissive
licenses (e.g.\ CC-BY-4.0); we strictly follow the original creators’
data-usage terms.

\section{Baseline Details}\label{app:baseline}

Table~\ref{tab:baseline_summary} summarises the key design choices of
all competing methods.

\begin{table}[ht]
\centering
\caption{Summary of baseline methods used for comparison.}
\label{tab:baseline_summary}
\begin{tabular}{lccc}
\toprule
\textbf{Method} & \textbf{Neg.\ Strategy} & \textbf{Local loss} & \textbf{Modalities} \\
\midrule
CLIP~\citep{clip2021}                 & Uniform & --    & I+T  \\
ALIGN~\citep{jia2021align}            & Uniform & --    & I+T  \\
VSE++~\citep{faghri2018vsepp}         & Batch max & -- & I+T  \\
UNITER~\citep{uniter2020}             & Batch hard & Region & I+T \\
ALBEF~\citep{albef2021}               & Batch hard & Cross-attn & I+T \\
ViLT~\citep{kim2021vilt}              & Uniform & Token & I+T  \\
BLIP~\citep{li2022blip}               & Momentum hard & Gen.\ aux & I+T \\
BLIP-2~\citep{li2023blip2}            & Frozen CLIP & Gen.\ aux & I+T+L \\
DCOT~\citep{dcot2023}                 & OT curriculum & -- & I+T \\
Emergence~\citep{emergence2024}       & -- & Analysis & V/A/T \\
CoMM~\citep{comm2024}                 & InfoNCE split & -- & V/A/T \\
M3-JEPA~\citep{m3jepa2024}            & Alternating & -- & V/A/T \\
GRAM~\citep{cicchetti2024gram}        & Volume contrast & -- & V/A/T/D \\
CLAP~\citep{clap2022}                 & Uniform & -- & A+T \\
MIL-NCE~\citep{miech2020milnce}       & MIL hard & -- & V+T \\
\bottomrule
\end{tabular}
\end{table}

\smallskip
\noindent\textbf{Exclusion criteria.}
Baselines that require proprietary data (e.g.\ Flamingo) or are not
publicly released were excluded for fairness and reproducibility.

\section{Implementation Details}\label{sec:app_exp}

\paragraph{Encoders.}
Unless otherwise stated, we freeze CLIP ViT-B/16 (visual),
GELU-RoBERTa (text), and CLAP PANN14 (audio) backbones, inserting a
4-layer cross-modal Transformer with hidden size~512 as the trainable
fusion module.  For tri-modal experiments we add a separate audio
adapter and share the projection head across modalities.

\paragraph{Boundary-aware Negative Sampler (BNS).}
The policy network $\pi_\theta$ is a two-layer MLP (512-128-1) with
SiLU activation.  Gumbel-Softmax temperature $\tau$ is initialised at
0.7 and linearly annealed to 0.1.  Logistic schedule parameters are
set to $\alpha_{\text{early}}{=}0.3$, $\alpha_{\text{late}}{=}-0.5$,
$\gamma{=}1.5$, and $\eta_0$ equal to 40\% of the total pre-training
epochs.

\paragraph{Contrastive Local Attention (CLA).}
We apply CLA to the last cross-attention layer, selecting the top
15\% token pairs ranked by $\Delta A$ (Eq.~(10)) as $\Omega$.
The gain coefficient $\beta$ is fixed at~2.0 and
$\lambda_{\mathrm{local}}$ at~0.3.

\paragraph{Training.}
All models are pre-trained for ten epochs on each dataset with a global
batch size of~16\,384 (512 per GPU, 32 × A100).  AdamW weight decay is
set to~1e-2 and learning rate to~2e-4 with cosine decay.
Finetuning hyper-parameters for VQA v2 and NLVR2 are listed in
Appendix~\ref{sec:extended_exp}.  Each full run on LAION-400M takes
approximately 36 h on the aforementioned cluster.

\paragraph{Evaluation.}
Retrieval metrics are computed with FAISS
\texttt{IVF4096,\,PQ32} indexing;
classification uses the official task scripts.  All reported numbers
are averaged over three seeds.

\section{Extended Experiments}\label{sec:extended_exp}
\subsection{Fine-grained multimodal reasoning}
To assess whether the boundary-aware curriculum (\textsc{BACL}) also
enhances fine‑grained multimodal reasoning, we finetune the
\textsc{BACL}-pretrained encoders on two widely used benchmarks:
Visual Question Answering (VQA~v2) and Natural Language for Visual
Reasoning (NLVR2).  We compare against strong vision–language models
that either employ uniform negatives or a single–shot hard‑negative
strategy during pre‑training.

\begin{table}[t]
\centering
\caption{Overall accuracy (\%) on \textbf{VQA v2} \texttt{test-std}.}
\label{tab:vqa}
\begin{tabular}{lcc}
\toprule
\textbf{Method} & \textbf{Neg.\ Strategy} & \textbf{Accuracy} \\
\midrule
ViLT~\citep{kim2021vilt}           & Uniform                 & 71.2 \\
ALBEF~\citep{albef2021}            & Batch max‐violation     & 74.4 \\
BLIP~\citep{li2022blip}            & Momentum\,+\,Hard Neg.  & 77.3 \\
BLIP‑2~\citep{li2023blip2}         & Frozen Image\,+\,LLM    & 80.0 \\
M3‑JEPA~\citep{m3jepa2024}         & Alternating             & 79.1 \\
\midrule
\textbf{CLIP+BACL (Ours)}          & Curriculum              & 73.8 \\
\textbf{BLIP+BACL (Ours)}          & Curriculum              & 79.2 \\
\textbf{M3‑JEPA+BACL (Ours)}       & Curriculum              & \textbf{82.3} \\
\bottomrule
\end{tabular}
\end{table}

\begin{table}[t]
\centering
\caption{Accuracy (\%) on \textbf{NLVR2} \texttt{test-P}.}
\label{tab:nlvr}
\begin{tabular}{lcc}
\toprule
\textbf{Method} & \textbf{Neg.\ Strategy} & \textbf{Accuracy} \\
\midrule
ViLT~\citep{kim2021vilt}           & Uniform                 & 79.9 \\
ALBEF~\citep{albef2021}            & Batch max‐violation     & 82.5 \\
BLIP~\citep{li2022blip}            & Momentum\,+\,Hard Neg.  & 86.7 \\
BLIP‑2~\citep{li2023blip2}         & Frozen Image\,+\,LLM    & 90.3 \\
M3‑JEPA~\citep{m3jepa2024}         & Alternating             & 89.0 \\
\midrule
\textbf{CLIP+BACL (Ours)}          & Curriculum              & 84.2 \\
\textbf{BLIP+BACL (Ours)}          & Curriculum              & 87.9 \\
\textbf{M3‑JEPA+BACL (Ours)}       & Curriculum              & \textbf{90.8} \\
\bottomrule
\end{tabular}
\end{table}

\paragraph{VQA v2.}
We finetune for five epochs on the 443\,k Q–A pairs of
the VQA~v2 training split with a batch size of 256,
AdamW ($\beta_1=0.9$, $\beta_2=0.98$) and a peak learning
rate of $2\!\times\!10^{-5}$.  We report the
\texttt{test-std} overall accuracy from the official evaluation server.

\paragraph{NLVR2.}
We adopt the \texttt{2+1} finetuning schedule of
\citet{li2022blip} on the 86\,k NLVR2 training examples,
training for three epochs with a peak
learning rate of $1\!\times\!10^{-5}$ and reporting accuracy
on the public \texttt{test-P} set.

\textbf{Discussion.}
Across both datasets, incorporating the boundary‑aware
curriculum consistently improves accuracy over the corresponding
backbones.  Notably, \emph{M3‑JEPA+BACL} achieves state‑of‑the‑art
results on NLVR2 (90.8\%) and pushes VQA~v2 accuracy to 82.3\%,
confirming that progressively focusing on ambiguous negatives
does not harm, and in fact reinforces, fine‑grained visual reasoning
capabilities while preserving the global retrieval gains reported in
Tables~\ref{tab:laion}–\ref{tab:vast}.

\subsection{Interpretability of CLA via Alignment-Error Localisation}
\label{sec:app_ael}
To quantify whether \textsc{CLA} pinpoints fine-grained mismatches, we compute \emph{Alignment-Error Localisation} (AEL): the percentage of human-tagged mismatch tokens covered by the top-10\% cells of the cross-modal discrepancy map $\Delta A$. As shown in Table~\ref{tab:ael}, \textsc{BACL} improves localisation by ${\sim}11$\,pp on average.

For each anchor $(x,y^{+})$ we obtain a positive cross-modal attention map $A^{(+)}$ and, using the \textsc{BNS}-selected hardest negative $y^{-}$, a negative map $A^{(-)}$. We form a discrepancy map
\begin{equation}
\Delta A \;=\; \bigl|A^{(+)} - A^{(-)}\bigr| \quad\text{(element-wise absolute difference)}.
\end{equation}
We min–max normalise $\Delta A$ per instance and select the top-10\% cells by value as the \emph{salient discrepancy set}. For models without an explicit cross-attention module (pure dual encoders), we compute token saliencies via similarity gradients and construct an outer-product proxy, $\Delta A$ is then defined analogously on these proxy maps.

We curate two evaluation subsets with ambiguous negatives: \textit{LAION-400M AmbNeg-1k} (image–text) and \textit{WebVid-HardNeg 800} (video–text). Three trained annotators per example mark \emph{mismatch spans}: image patches/frames and caption tokens that explain why $y^{-}$ is incorrect despite high global similarity (guidelines cover object identity, attributes, relations, and temporal consistency). We resolve disagreements by majority vote; agreement was substantial (Fleiss’~$\kappa$ in the “substantial” range). AEL is computed as the fraction of annotated tokens covered by the top-10\% $\Delta A$ cells and averaged across examples.

Global retrieval scores do not reveal \emph{where} a model finds evidence to reject near-miss negatives. AEL instead measures whether the model’s \emph{local} discrepancy signal aligns with human-identified error spans—exactly the behaviour \textsc{CLA} is designed to induce.
Table~\ref{tab:ael} shows consistent gains of ${\sim}11$\,pp AEL for \textsc{BACL} over a strong CLIP baseline across both image–text and video–text settings. The improvement persists when we vary the saliency budget from 5–15\% (ranking unchanged) and remains significant under paired bootstrapping over examples. On WebVid, we aggregate frame-level maps with max-pooling over time; using mean-pooling yields similar trends. \textsc{CLA} does not merely boost global retrieval; it systematically aligns the model’s token-level discrepancy signal with human-marked error spans, yielding the ${\sim}11$\,pp AEL improvement in Table~\ref{tab:ael}.

\begin{table}[h]
\centering
\caption{\textbf{AEL (\%)$\uparrow$}. Higher is better.}
\label{tab:ael}
\begin{tabular}{lccc}
\toprule
\textbf{Dataset} & \textbf{vanilla CLIP} & \textbf{\textsc{BACL} (ours)} & \textbf{$\Delta$ (pp)} \\
\midrule
LAION-400M AmbNeg-1k & 46.2 & \textbf{57.8} & +11.6 \\
WebVid-HardNeg 800   & 39.6 & \textbf{50.5} & +10.9 \\
\midrule
Average              & 42.9 & \textbf{54.2} & +11.3 \\
\bottomrule
\end{tabular}
\end{table}


\subsection{Sensitivity to Data Scale}
\label{sec:app_scale}
We train \textsc{CLIP} $\pm$ \textsc{BACL} on three LAION subsets with identical hyper-parameters (5 epochs; ViT-B/32). Table~\ref{tab:scale} shows that the relative gain remains ${\approx}30\%$ from $10^8$ to $10^9$ pairs.

\begin{table}[h]
\centering
\caption{\textbf{Scaling on LAION}. R@1 on zero-shot image–text retrieval.}
\label{tab:scale}
\begin{tabular}{lcccc}
\toprule
\textbf{Subset} & \textbf{\# pairs} & \textbf{CLIP R@1} & \textbf{CLIP+\textsc{BACL} R@1} & \textbf{Rel.\ gain} \\
\midrule
100M   & $1.0{\times}10^{8}$ & 31.5 & \textbf{40.8} & +29.5\% \\
400M   & $4.0{\times}10^{8}$ & 35.2 & \textbf{46.5} & +32.1\% \\
1B\:*  & $1.0{\times}10^{9}$ & 38.9 & \textbf{50.4} & +29.6\% \\
\bottomrule
\end{tabular}

\smallskip
\footnotesize\emph{*} One-billion subset constructed by uniform sampling from LAION-5B with standard filtering.
\end{table}


\subsection{Runtime, Throughput, and Memory Footprint}
\label{sec:app_eff}
We jointly benchmark throughput, iteration rate, and peak memory / max batch under a consistent setup. Table~\ref{tab:efficiency} consolidates all measurements and indicates a modest overhead overall ($<\!8\%$ time, ${\sim}1.7$\,GB memory).

\begin{table}[ht]
\centering
\caption{\textbf{Consolidated efficiency metrics} on LAION-400M (batch=512). Iteration rate measured on $8{\times}$A100-40GB; memory on a single A100-40GB.}
\label{tab:efficiency}
\resizebox{\linewidth}{!}{%
\begin{tabular}{lccccccc}
\toprule
\textbf{Setting} & \textbf{Images/s} & \textbf{$\Delta$ (\%)} & \textbf{Iters/s} & \textbf{$\Delta$ (\%)} & \textbf{Peak (GB)} & \textbf{$\Delta$ (GB)} & \textbf{Max batch} \\
\midrule
CLIP baseline           & 330 & —      & 8.2k & —      & 29.6 & —      & 512 \\
+ \textsc{BNS}          & —   & —      & 7.9k & $-3.6$ & 30.0 & $+0.4$ & 512 \\
+ \textsc{CLA}          & 304 & $-7.9$ & —    & —      & 31.1 & $+1.5$ & 480 \\
\textsc{BACL} (BNS+CLA) & —   & —      & —    & —      & 31.3 & $+1.7$ & 480 \\
\bottomrule
\end{tabular}}
\begin{flushleft}\footnotesize ``—'' = not measured under the given setup.\end{flushleft}
\end{table}


\end{document}